\documentclass[runningheads]{llncs}

\usepackage{eccv}

\usepackage{eccvabbrv}

\usepackage{graphicx}
\usepackage{booktabs}

\usepackage[accsupp]{axessibility}  

\usepackage{hyperref}

\usepackage{orcidlink}

\usepackage[dvipsnames]{xcolor}
\definecolor{LB}{HTML}{ECF4FF}

\usepackage{bm}

\DeclareMathOperator*{\argmin}{arg\,min}
\usepackage{multirow}
\usepackage{colortbl}
\usepackage{utfsym}
\usepackage{wrapfig}
\usepackage{enumitem}

\begin{document}


\newcommand{\htcomment}[1]{\begin{center}\fbox{\begin{minipage}{.4\textwidth}HT: #1\end{minipage}}\end{center}}

\newcommand{\slim}{\emph{\small SLIM}}
\newcommand{\slimTr}{\emph{\small SLIM$_{Tr}$}}
\newcommand{\slimVal}{\emph{\small SLIM$_{Val}$}}

\title{\emph{SLIM}: Spuriousness Mitigation\\with Minimal Human Annotations} 


\titlerunning{\emph{SLIM}: Spuriousness Mitigation with Minimal Human Annotations}

\author{Xiwei Xuan\inst{1}\orcidlink{0000-0002-0828-8761} \and
Ziquan Deng\inst{1}\orcidlink{0000-0003-1548-5197} \and
Hsuan-Tien Lin\inst{2}\orcidlink{0000-0003-2968-0671} \and
Kwan-Liu Ma\inst{1}\orcidlink{0000-0001-8086-0366}}

\authorrunning{X.~Xuan et al.}

\institute{University of California, Davis \and National Taiwan University\\
\email{\{xwxuan, ziqdeng, klma\}@ucdavis.edu, htlin@csie.ntu.edu.tw}}

\maketitle

\begin{abstract}
Recent studies highlight that deep learning models often learn spurious features mistakenly linked to labels, compromising their reliability in real-world scenarios where such correlations do not hold. Despite the increasing research effort, existing solutions often face two main challenges: they either demand substantial annotations of spurious attributes, or they yield less competitive outcomes with expensive training when additional annotations are absent. In this paper, we introduce \slim, a cost-effective and performance-targeted approach to reducing spurious correlations in deep learning. Our method leverages a human-in-the-loop protocol featuring a novel attention labeling mechanism with a constructed attention representation space. \slim~significantly reduces the need for exhaustive additional labeling, requiring human input for fewer than $3\%$ of instances. By prioritizing data quality over complicated training strategies, \slim~curates a smaller yet more feature-balanced data subset, fostering the development of spuriousness-robust models. Experimental validations across key benchmarks demonstrate that \slim~competes with or exceeds the performance of leading methods while significantly reducing costs. The \slim~framework thus presents a promising path for developing reliable models more efficiently.
Our code is available in \url{https://github.com/xiweix/SLIM.git/}.


\end{abstract}

\section{Introduction}\label{sec:intro}
\begin{figure*}[tbhp]
  \centering
   \includegraphics[width=1\linewidth]{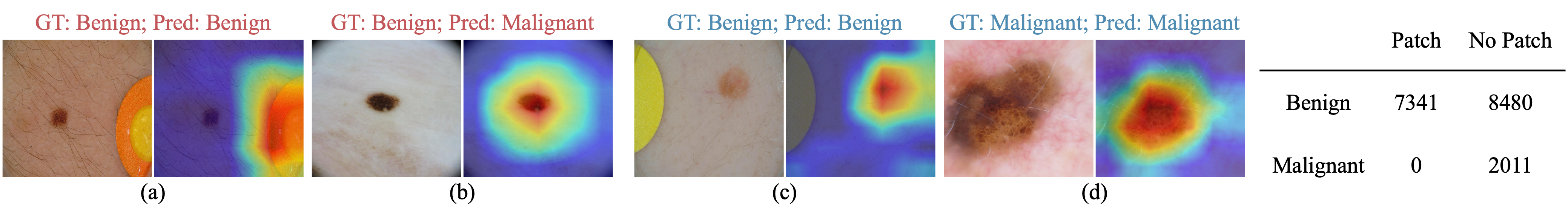}
\caption{(Left) {\small GradCAM} visualizations highlighting model attention on {\small ISIC}. {\small (a, b)} showcase {\color{BrickRed} a model biased towards patches}, leading to a correct prediction with wrong reasons in (a) but an incorrect prediction despite correct focus in (b). {\small (c, d)} depict {\color{MidnightBlue} a spuriousness-robust model} consistently focusing on core features. (Right) The table details the training data distribution by class (benign or malignant) and the presence of color patches, illustrating the imbalance and potential for spurious correlations.
}
   \label{fig:intro}
\end{figure*}

Spurious correlations, where models mistakenly rely on irrelevant features to make decisions, pose a significant challenge in machine learning.
Consequently, certain groups may experience deceptively inflated performance metrics, while accuracy reduces for others. This prevalence of ``right answers for wrong reasons'' fosters a misleading impression of model proficiency and constrains its utility across diverse contexts.
Meanwhile, rectifying the issue of spuriousness is a challenging task, particularly when seeking to enhance the performance of underperforming groups.
A notable example is the {\small ISIC} skin cancer dataset\cite{8363547}, where a typical spurious correlation occurs when color patches frequently co-occur with benign-labeled data. As shown in Fig.~\ref{fig:intro}, a biased model may focus on patches rather than skin lesions to identify melanoma, and fail when such patches are missing, while a desirable case would be the model captures correct skin regions regardless of spurious features’ existence. 
Enhancing model robustness to address such dangerous misdiagnoses is crucial, driving a surge in research towards more effective solutions.

\setlength{\floatsep}{2pt plus 2pt minus 2pt}
\setlength{\textfloatsep}{2pt plus 2pt minus 2pt}
\setlength{\intextsep}{2pt plus 2pt minus 2pt}
\setlength{\floatsep}{2pt plus 2pt minus 2pt}
\setlength{\textfloatsep}{2pt plus 2pt minus 2pt}
\setlength{\intextsep}{2pt plus 2pt minus 2pt}
\begin{wraptable}{r}{0.46\textwidth}
\centering
\caption{Annotation forms and quantities required by different methods on {\small Waterbirds}, highlighting the reduced workload of \slim. {\small (\textit{Att.=Attention, T.=Training set, V.=Validation set})}}
\label{tab:anno_form_compare}
\resizebox{0.45\textwidth}{!}{%
\begin{tabular}{p{3cm}<{\centering}p{3cm}<{\centering}p{3cm}<{\centering}}
Method             & Anno. Form  & Anno. Amount \\ \hline
RRR\cite{ijcai2017p371}                & Map              & $100\% \times$ T.              \\
GradIA\cite{10.1145/3555590}             & Att. Yes-No, Map      & $100\% \times$ T.              \\
Energy Loss\cite{rao2023using}        & Bounding box     & $100\% \times$ T.              \\
CRAYON\cite{leetowards} & Att. Yes-No           & $100\% \times$ T.              \\
GDRO\cite{Sagawa2020Distributionally}               & Spuriousness label      & $100\% \times$ T.              \\
DFR\cite{kirichenko2022last}                & Spuriousness label        & $100\% \times$ V.              \\
\rowcolor[HTML]{EFEFEF} 
\slim~(ours)               & Att. Yes-No           & $2.5\% \times$T. or $3\% \times$V.                \\
\end{tabular}%
}
\end{wraptable}

A typical approach in this context is to seek for optimal data distribution. {\small GDRO}\cite{Sagawa2020Distributionally} proves the efficacy of group-balanced data in enhancing model robustness. However, along with subsequent studies\cite{nam2022spread,kirichenko2022last}, such an approach requires comprehensive spuriousness labels at a dataset scale, consuming a considerable cost. Our investigations reveal notable inconsistencies in human-annotated spuriousness labels, raising concerns about their reliability (refer to Table~\ref{tab:human_annotation_eval}). There are other approaches with more direct human involvement, such as providing yes-no feedback on attention correctness\cite{xuan2024attributionscanner,leetowards}; or outlining core features with maps\cite{ijcai2017p371,10.1145/3555590} or bounding boxes\cite{rao2023using}. Despite different formats of queried information, they require extensive effort across the entire dataset, as summarized in Table~\ref{tab:anno_form_compare}. Considering the empirical applicability, our work aims to streamline this process by reducing human annotation workload and minimizing the ambiguity of collected data.

Methods requiring spuriousness annotation tend to excel in performance at higher costs, while minimizing such expenses typically results in performance trade-offs\cite{creager2021environment,ahmed2020systematic}. Facing this challenge, techniques such as {\small JTT\cite{liu2021just}} and {\small CNC\cite{zhang2022correct}} design training or data-augmenting phases to reduce spuriousness without additional annotations. However, the extra computational cost raised by multiple training iterations is less desirable. We believe, as spuriousness is indeed caused by data biases, improving data quality is more essential than designing complex learning strategies.

Therefore, the remaining research gap calls for a data quality-oriented approach that balances performance and cost. This is where our framework, \slim, steps in. \slim~offers a novel approach to mitigating spurious correlations by intelligently leveraging a minimal set of human-annotated data and curating a feature-balanced data set to enhance model robustness without any significant computational overhead incurred by many existing methods.

As shown in Fig.~\ref{fig:framework}, inferred from a reference model, \slim~first constructs an attention representation space that reflects locally consistent data features and model attention, enabling efficient attention label estimation. --- Representative instances can be sampled from this space for querying attention correctness labels and such labels can be expanded to neighbor instances. Subsequently, utilizing a visual explanation method that highlights model attention, like {\small GradCAM}\cite{selvaraju2017grad}, \slim~disentangles core and environment features within the latent space. Here, ``core features'' denote features critical for class determination, whereas ``environment features'' represent the surrounding contextual information. We agree with {\small GDRO} that balanced data distribution is crucial, but instead of using human-annotated labels to approximate spuriousness distribution, we leverage latent space features directly disentangled from model attention to balance data distribution, which is more aligned with model understanding and involving less human assumptions. Through this, \slim~assembles data groups with diverse feature combinations. Focused on data quality, a consistency-aware sampling mechanism is employed to curate a subset with optimal feature distributions — ensuring balanced core features across various environmental contexts. Finally, we conclude our framework with a straightforward training procedure.

The experimental results demonstrate \slim's advanced capability in spuriousness mitigation by performing comparably to, or exceeding state-of-the-art (SOTA) methods. Notably, \slim~accomplishes this with (1) a significantly reduced requirement for human annotations: $0.12\%$-$2.5\%$ of the training set size and $0.35\%$-$4\%$ of the validation set; and (2) a lightweight training for a robust model using our constructed data with $5\%$-$30\%$ of the original set. 
Our code will be published after the review process is complete. Our contributions are:
\begin{itemize}[leftmargin=*]
\item \textbf{{\small Data-quality-oriented spuriousness mitigation:}} \slim~introduces an interactive, simple, and effective data construction pipeline to create high-quality data for spuriousness mitigation.
\item \textbf{{\small Cost-efficient framework:}} In contrast to related works with leading performance, \slim~demonstrates a significant reduction in both annotation requirements and computational cost.
\item \textbf{{\small Competitive or superior model performance:}} Without incurring excessive costs, \slim~achieves comparable or superior performance to existing methods, confirming its effectiveness in mitigating spurious correlations.
\end{itemize}

\section{Related Work}\label{sec:related}
\noindent\textbf{Spuriousness mitigation with spuriousness labels.} 
When a comprehensive set of spuriousness labels is accessible during training, methods like class balancing\cite{he2009learning} and importance weighting\cite{taghanaki2021robust} can be employed for mitigating spurious correlations, which utilizes data groups defined by class and spuriousness labels. Following the same schema, GDRO\cite{Sagawa2020Distributionally} refines the training approach by focusing on the worst-performing groups as determined by ERM loss.
Aiming to reduce the dependency on comprehensive spuriousness labeling, several methods\cite{sohoni2020no,nam2020learning,liu2021just,zhang2022correct,creager2021environment,ahmed2020systematic,taghanaki2021robust} have emerged to approximate group information using a reference model. While these approaches are technically sophisticated, they often fail to achieve the same level of performance as approaches informed by spuriousness labels.
To balance the tradeoff between performance and cost, a further research direction reduces the requirement for extensive labeling by leveraging spuriousness labels in smaller validation datasets\cite{nam2022spread,kirichenko2022last}. For example, SSA\cite{nam2022spread} adopts semi-supervised learning to predict group labels in training data according to information from validation, then applies GDRO to reduce spurious correlations. However, even with this reduced scale of validation sets, the labor-intensive nature of spuriousness labeling remains a barrier to scalability. Moreover, groups in these methods are formed based on human-defined criteria, risking a misalignment with the actual categories a model might discern. This drawback can hinder the true representation of spurious correlations and may lead to suboptimal mitigation outcomes.

\noindent\textbf{Spuriousness mitigation with human-in-the-loop.}
Integrating human insights has proven beneficial in mitigating model spuriousness, as evidenced by various human-involved strategies. Considering the valuable human effort, \cite{schramowski2020making} provides an efficient solution by querying data samples for issue mitigation with domain experts in the loop, which overlooks potential issues in the broader data scope.
Another option for avoiding redundant human annotation is to query global spurious concepts\cite{srivastava2020robustness,yan2023towards} from knowledgeable experts, which can be ambiguous and fail to cover all possibilities.
More recent approaches\cite{ijcai2017p371,gao2022aligning,rao2023using} raise concerns about the inconsistency of spuriousness labels, introducing alternatives such as human-generated attention maps or bounding boxes outlining core features. These human-generated references help align the models' attention with the actual features relevant to decision-making. However, producing such annotations is challenging, and these techniques are resource-intensive as they necessitate annotations across the entire dataset.
In response, more recent research research\cite{bontempelli2023conceptlevel,10.5555/3618408.3620051} simplifies the annotation process to binary feedback on model's attention correctness. While less demanding, its requirement to annotate all instances still poses challenges in real practices.

\section{Problem Formulation}\label{sec:problem}
We introduce the formal definition of the spuriousness mitigation problem\cite{joshi2023towards}. Assume training data $\mathcal D=\{(\bm{x}_i,y_i)\}_{i=1}^n$ with input features $\bm{x}_i\in \mathbb{R}^d$ and labels $y_i\in [\mathcal L]$, where $\mathcal L= \{L_1,\dots,L_l\}$ are the classes in $\mathcal D$.
Machine learning models for classification are trained to minimize the empirical risk:

\begin{equation}\label{eqn:obj}
    \bm{\theta}^*\in \mathop{\argmin}\limits_{\bm{\theta}} \mathop{\mathbb{E}}\limits_{(\bm{x_i},y_i)\in\mathcal D}[l(f(\bm{\theta},\bm{x}_i),y_i)],
\end{equation}
where $\bm{\theta}$ represent model parameters, $f(\bm{\theta},\bm{x}_i)$ and $l(*)$ are the model outputs and the loss function, respectively.

\noindent\textbf{Groups.}
In spuriousness mitigation, we assume there exists a core feature set $\mathcal C= \{c_1,\dots,c_l\}$ and a spurious feature set $\mathcal S= \{s_1,\dots,s_k\}$ in $\mathcal D$. Each class $L_i$ is uniquely identified by a core feature $c_i$, while spurious features $s_j$ can be found across different classes. We consider an instance in the dataset as having both a core feature $c_i$ and a spurious feature $s_j$, and each instance belongs to a unique group $g_{c_i, s_j}$.
Core features are reliable class indicators in both training and testing sets, while spurious features with misleading co-occurrence with class in training may miss such correlations in the testing phase. If the model mistakenly links spurious features with classes in training, it may fail in testing when such spurious features are absent. This can lead to lower accuracy rates, especially for less common group variations that do not exhibit the spurious feature the model has learned to rely on.

\noindent\textbf{Worst-group accuracy.}
The vital evaluation of spuriousness mitigation focuses on the model's performance in the subgroup where it is least effective -- the worst-group accuracy. It is calculated by:
\begin{equation}\label{eqn:worst_group_acc}
Acc_{w} = \min_{g \in \mathcal G} ( \frac{1}{|g|} \textstyle\sum_{(x_i, y_i) \in g} \mathbf{1} [f(x_i) = y_i] ),
\end{equation}
where $\mathcal G = \{\cup_{c_i,s_j} g_{c_i,s_j}\}$ is the union of subgroups defined by core feature $c_i$ and spurious feature $s_j$, and $f(x_i)$ is the label predicted by the model.
This metric is crucial because it directly assesses how effectively a model can generalize across different subgroups, especially those that might be mostly underrepresented because of spurious correlations. It ensures that improvements in model performance are not just reflective of gains on the majority or less challenging subsets of the data.

\begin{figure*}[tbhp]
  \centering
   \includegraphics[width=0.95\linewidth]{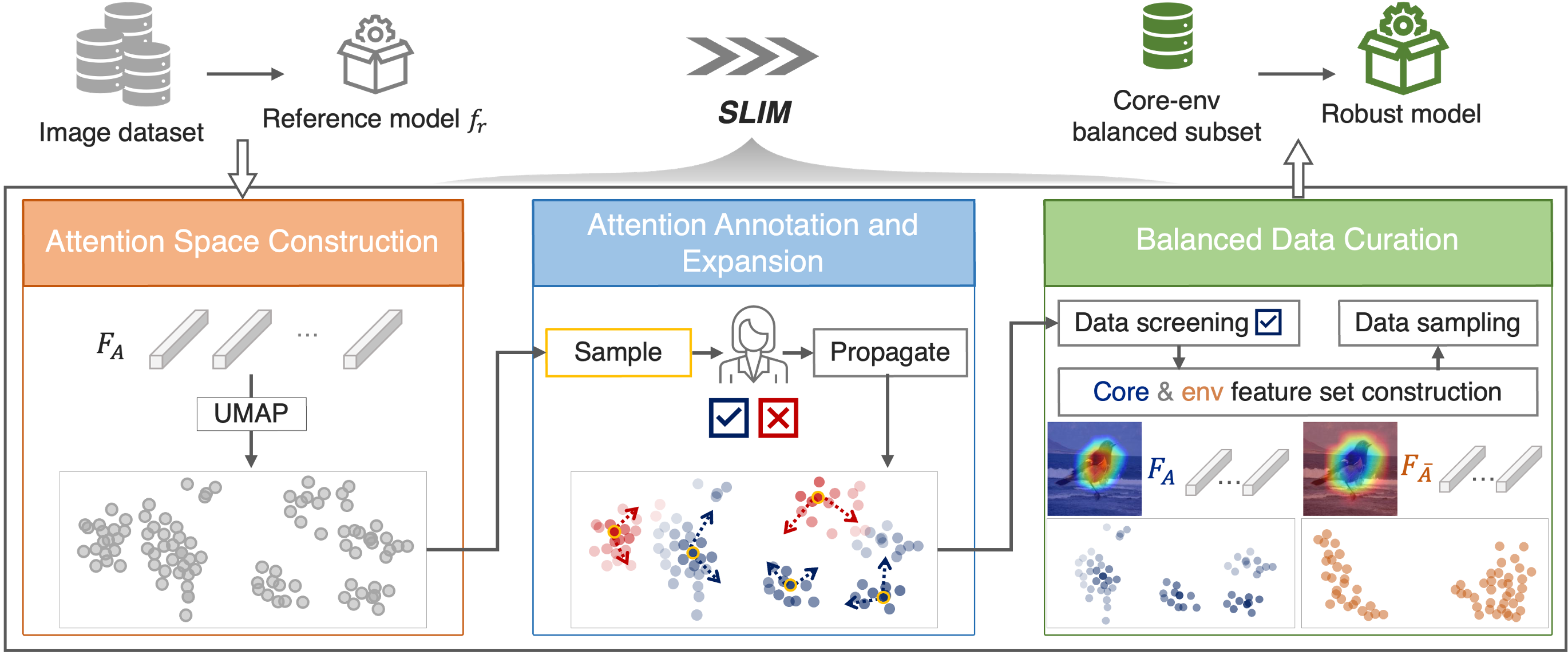}
\caption{
Overview of \slim~framework, with a data construction pipeline consisting of three phases: (1) {\small \textbf{Attention Space Construction}}, creating a space with data features and model attention aligned locally; (2) {\small \textbf{Attention Annotation and Expansion}}, where instances sampled from the attention space are labeled by human for attention correctness, and labels are propagated to neighboring instances; and (3) {\small \textbf{Balanced Data Curation}}, which filters out instances with incorrect attention, and utilize attention-weighted ($F_A$) and inverse-attention-weighted ($F_{\bar{A}}$) feature vectors to create core and environment feature sets, forming subgroups to assemble a feature-balanced subset for training a spuriousness-robust model.
}
   \label{fig:framework}
\end{figure*}

\section{Method}\label{sec:method}
In this section, we detail the method behind~\slim. 
In alignment with related works requiring no spuriousness label, our approach also follows a two-stage procedure: inferring groups with an ERM-trained reference model $f_r$, and then leveraging the inferred groups to train a robust model.
As an approach oriented by data quality, \slim~framework comprises a three-phase data construction pipeline, as illustrated in Fig.~\ref{fig:framework}. Firstly, we build an attention representation space to group instances with both similar data features and model attention, preparing for the data sampling and label propagation. Following this, we sample typical instances for human to evaluate model's attention correctness with yes-no feedback, then estimate neighboring instances' attention labels accordingly. In the last phase, we filter out data with wrong attention and leverage attention matrix to disentangle core and environment features. With constructed feature sets, we apply consistency-weighted data sampling to curate a feature-balanced subset, with a desired core-environment balanced distribution. Finally, \slim~framework is concluded by a training process with the curated subset, resulting in a model exhibiting robustness against spurious correlations.

\subsection{Attention Space Construction}
\label{subsec:att_rep_space_construction}

Considering the benefits of being aware of spuriousness information and drawbacks of the annotation cost, in our framework, we deploy a binary labeling scheme to simplify descriptive labeling tasks into yes-no feedback on model's attention correctness. Moreover, we aim to further facilitate this process by avoiding the necessity of exhaustive labeling. Our empirical observations indicate that a model would have consistent attention mechanisms on specific data subgroups. If such subgroups can be found, it would largely aid the labeling workload by enabling label estimations among neighbors.

Driven by this, we aim to find a representation space where instances with similar attention outcomes (as indicated by highlighted image features) are closely grouped. 
However, our analysis reveals that conventional feature representation spaces formed directly from feature vectors do not fulfill this requirement -- such spaces only maintain local consistency in data features but fail to capture the nuances of model attention (refer to Sec.~\ref{subsec:att_eval_performance}).

\begin{figure*}[bthp]
  \centering
   \includegraphics[width=0.95\linewidth]{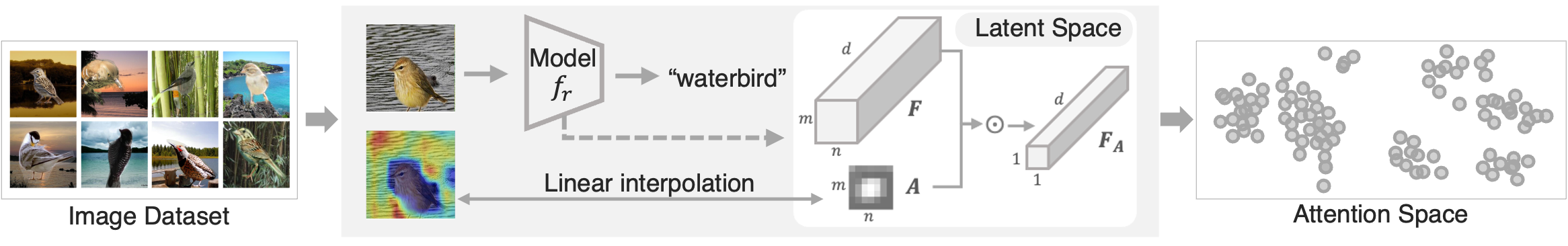}
\caption{
Construction of the attention space. In the latent space, feature vector $F$ and model's attribution vector $A$ are extracted, representing input features and model's attention, representatively. By weighting $F$ with $A$, the attention-weighted feature vector $F_A$ emphasizes features of model's top interest. All $F_A$ are then projected to form the attention space featuring locally consistent features and model attention.
}\label{fig:att_space_construction}
\end{figure*}
Thus, we have developed a process utilizing model attention in the latent space to support the establishment of such space, as illustrated in Fig.~\ref{fig:att_space_construction}.
For each image, we obtain its feature vector $F$ and model attribution $A$ in the latent space at first. Note that $A$ represents the model's attention focus and can be linearly interpolated to visualize an attention heatmap.
We then use $A$ as a weight matrix for $F$, producing an attention-weighted feature vector $F_A$. 
Finally, with the dimensionality reduction method UMAP\cite{mcinnes2018umap}, we project vectors $F_A$ into a 2D space, namely the attention space, which ensures local consistency encompasses both data features and the model's attention mechanisms.

Note that for a fair comparison, in alignment with baseline methods, experiments in this paper involve the CNN model (ResNet) for classification and GradCAM for obtaining model attention. But our framework is fully applicable to (1) other vision model architectures such as vision transformers (ViT)\cite{dosovitskiy2020image} or Swin transformers\cite{liu2021swin}, and (2) other heatmap-based model visualization techniques, such as RISE\cite{petsiuk2018rise} or ScoreCAM\cite{wang2020score}.

\subsection{Attention Annotation and Expansion}
\label{subsec:att_eval_propagate}

Given that our attention space ensures locally consistent model attention, it allows for annotating a selectively sampled subset of instances and then estimating neighboring labels accordingly, thereby enabling efficient transfer of human expertise. The criteria for selecting these instances are twofold: (1) \textit{diversity}, to encompass a broad covering of various data features and model attention patterns; (2) \textit{typicality}, ensuring they are representative of their neighboring instances. To achieve this, we utilize the k-means clustering algorithm to partition our constructed attention space into $n$ clusters (\textit{diversity}). We then select $n$ instances that are closest to each cluster's center (\textit{typically}). The choice of $n$ is aimed at optimizing the balance between minimizing intra-cluster variance in data features and model attention patterns, and keeping the annotation budget manageable. Our empirical analysis, detailed in Sec.~\ref{subsec:att_eval_performance}, reveals that the intra-cluster consistencies increase with $n$, but this trend levels off upon reaching a specific value of $n$--the elbow point.
Thus, selecting $n$ at this elbow point optimizes consistency without unnecessarily increasing the annotation workload.

We gather binary yes-no feedback from human experts on the correctness of the model's attention for these $n$ instances, feeding this feedback along with our attention space embedding into the label spreading algorithm. This algorithm leverages data similarity and density to extend attention labels to previously unlabeled data\cite{zhou2003learning}. Ultimately, this process yields a probability score for correct attention for each instance across the dataset.

\subsection{Balanced Data Curation}
\label{subsec:balance_data_construction}
Following Sec.~\ref{subsec:att_eval_propagate} that provides attention accuracy scores, we proceed to construct our data subset to support model's unbiased learning, as described below:

\noindent\textbf{Data screening.}
Inspired by works\cite{chen2022towards,jelassi2022towards,deng2023robust},
our theoretical analysis further supports the premise that data receiving correct attention from a reference model are more likely to enhance the learning of core features in a robust model within a more balanced data distribution, as substantiated in Appx.~A.2.
Furthermore, our study indicates that even a reference model with inherent biases has the potential to accurately identify core features for a data subset. Such correctly captured features typically exhibit diversity, ensuring good coverage across the spectrum of the core feature set. This understanding informs our approach to curating a data subset that: (1) selects instances based on accurate model attention, and (2) ensures a balanced representation of core features across various environment features.
Thus we first apply data screening to filter out instances with wrong attention according to attention accuracy scores.

\noindent\textbf{Disentangling core and environment features.}
After screening, the retained data subset mainly comprises instances with correct attention. This is reflected by attention masks that highlight core features, meaning that the corresponding attribution vector $A$ in the latent space also highlights core features of feature vector $F$. Moreover, existing studies have found that a model's feature vectors are representative of most input features, regardless of whether they are used for decision-making\cite{kirichenko2022last,xue2023eliminating}. Therefore, utilizing $A$ as the weight matrix, we disentangle core and environment features in the latent space --- weighting $F$ by attention masks $A$ and inverse attention masks $\bar{A}$. Subsequently, k-means clustering is applied to partition all attention-weighted feature vectors $F_A$, and inverse-attention-weighted feature vectors $F_{\bar{A}}$, respectively, facilitating the formation of representative subgroups for the core feature set ($\hat{\mathcal C}= {\hat{c}_1, \dots, \hat{c}_N}$) and the environment feature set ($\mathcal E= {e_1, \dots, e_M}$). The number of clusters, determined through the elbow method, culminates in $M \times N$ subgroups $g_{\hat{c}_i,e_j}$, each characterized by a distinct combination of core and environment features ($\hat{c}_i,e_j$).

\noindent\textbf{Data sampling.}
Given the unequal distribution of core and environment features, the internal consistency across each constructed subgroup $g_{\hat{c}_i,e_j}$ varies. Our sampling strategy aims to mitigate over-representation of particular features to prevent overfitting, while ensuring broad coverage of diverse core and environment features. We assess cluster consistency as a metric for determining the sample size from each cluster. For a cluster $\hat{c}_i$ in the core feature space, denote the distance between an instance $x$ and the cluster center as $d_{\hat{c}_i}(x)$. The cluster consistency is defined as the reciprocal of the average distance per cluster:
\begin{equation}\label{eqn:rho}
    \rho_{\hat{c}_i}= \frac{1}{\frac{1}{|\hat{c}_i|}\sum_{x\in\hat{c}_i}d_{\hat{c}_i}(x)}.
\end{equation}
A lower $\rho_{\hat{c}_i}$ signals less consistent features, necessitating a higher sample weight for adequately representing them. A cluster's sampling weight is defined as $w_{\hat{c}_i}=\frac{1}{\rho_{\hat{c}_i}}$ and the number of samples to be drawn from a core cluster $\hat{c}_i$ is:
\begin{equation}\label{eqn:n_c}
\begin{split}
     n_{\hat{c}_i}
     = N \cdot \frac{1/\rho_{\hat{c}_i}}{\sum_{j=1}^{|\hat{\mathcal C}|}1/\rho_{\hat{c}_j}},
\end{split}
\end{equation}
where $N$ is the sampling budget. Moving to the environment feature space, the $n_{\hat{c}_i}$ instances from a core feature cluster ${\hat{c}_i}$ may distribute across multiple environment clusters $\{e_1, \dots,  e_M\}$ and formulate subgroups $\{g_{\hat{c}_i, e_1}, \dots,  g_{\hat{c}_i, e_M}\}$. A similar methodology is employed for a certain group $g_{\hat{c}_i, e_i}$. The average distance for $g_{\hat{c}_i, e_i}$ in the environment representation space is:
\begin{equation}\label{eqn:rho2}
    \rho_{\hat{c}_i, e_i}= \frac{1}{\frac{1}{|\hat{c}_i, e_i|}\sum_{x\in(\hat{c}_i, e_i)}d_{e_i}(x)}.
\end{equation}
Similarly, we have the sampling weight $w_{\hat{c}_i, e_i}=\frac{1}{\rho_{\hat{c}_i, e_i}}$ and the number of samples to be drawn from a certain group $g_{\hat{c}_i, e_i}$ is finally defined by:
\begin{equation}\label{eqn:n_group}
\begin{split}
     n_{\hat{c}_i,e_i} &= n_{\hat{c}_i} \cdot \frac{w_{\hat{c}_i, e_i}}{\sum_{j=1}^{|\mathcal E(\hat{c}_i)|}w_{\hat{c}_j, e_i}}
     = N \cdot \frac{1/\rho_{\hat{c}_i}}{\sum_{j=1}^{|\hat{\mathcal C}|}1/\rho_{\hat{c}_j}} \cdot \frac{1/\rho_{\hat{c}_i, e_i}}{\sum_{j=1}^{|\mathcal E(\hat{c}_i)|}1/\rho_{\hat{c}_i, e_j}},
\end{split}
\end{equation}
where $|\hat{\mathcal C}|$ denotes the number of core clusters and $|\mathcal E(\hat{c}_i)|$ denotes the number of subgroups formulated by instances from cluster $\hat{c}_i$. To summary, our sampling strategy takes feature consistencies and sampling budget into consideration, and ensures fair representation of different core-environment feature combinations across the feature spaces.

\subsection{Training}
\label{subsec:training}

In our pursuit to mitigate spurious correlations, we emphasize the pivotal role of data quality by ensuring a balanced distribution of core features across various environmental contexts within our dataset. Given this focus, we employ a basic training approach for our model to learn from this well-prepared dataset.

We apply our data curation pipeline separately to the training (\slimTr) and validation (\slimVal) sets. For the data constructed from the training set, we adopt the conventional {\small ERM}. For the dataset derived from the validation set, we employ the training technique presented by {\small DFR}\cite{kirichenko2022last}. It involves retraining only the final linear layer of a classification model that was initially trained using ERM. Note that our approach diverges from {\small DFR} practices by utilizing our curated dataset instead of a metagroup-balanced dataset.

\section{Experiments}\label{sec:exp}


\subsection{Experimental Setup}
\label{subsec:experimental_setup}

\noindent\textbf{Datasets.}
Our study explores a range of datasets with inherent spurious correlations to challenge image classification models, including:
(1) {\small \textbf{Waterbirds}\cite{Sagawa2020Distributionally}.} Contrasting landbirds and waterbirds against congruent/incongruent backgrounds, this dataset tests models' ability to focus on relevant features for bird type classification.
(2) {\small \textbf{CelebA}\cite{liu2015deep}.} A comprehensive celebrity image dataset where the focus is on distinguishing hair color (blond vs. non-blond) to examine biases in associating hair color with gender.
(3) {\small \textbf{ISIC}\cite{8363547}.} Aiming to differentiate benign from malignant melanoma in dermoscopic images, we target colorful patches as spurious features, aligning with previous studies\cite{10.5555/3524938.3525689}.
(4) {\small \textbf{NICO}.} Derived from NICO++\cite{zhang2022nico}, this dataset features various object categories in shifted contexts to probe spurious correlations.
(5) {\small \textbf{ImageNet-9}\cite{xiao2020noise}.} Comprising 9 super-classes from {\small ImageNet}, this dataset is crafted to test models' robustness against background variations. Detailed information is available in Appx.~A.4.

\noindent\textbf{Baselines.}
We benchmark \slim~against a spectrum of SOTA spuriousness mitigation techniques, categorized as follows:
(1) Techniques not reliant on spuriousness labels: {\small EIIL}\cite{creager2021environment}, {\small PGI}\cite{ahmed2020systematic}, {\small GEORGE}\cite{sohoni2020no}, {\small LfF}\cite{nam2020learning}, {\small CIM}\cite{taghanaki2021robust}, {\small JTT}\cite{liu2021just}, and {\small CNC}\cite{zhang2022correct};
(2) Techniques requiring additional information such as predefined spuriousness concepts: {\small DISC}\cite{10.5555/3618408.3619982};
(3) Methods utilizing spuriousness labels from validation sets: {\small SSA}\cite{nam2022spread} and {\small DFR}\cite{kirichenko2022last};
(4) Approaches utilizing spuriousness labels from training sets: {\small GDRO}\cite{Sagawa2020Distributionally} and {\small PDE}\cite{deng2023robust}.

\noindent\textbf{Evaluation.}
Aligning with baselines, our experiments involve CNN models with ResNet50 architecture and GradCAM for visualizing model attention.
For the capability of performance enhancement, we evaluate the worst-group and average accuracy. 
Additionally, we consider the necessity for extra labels and the additional (add.) training cost to evaluate computational efficiency. 
Note that the add. training cost is aligned with\cite{joshi2023towards,yang2023identifying}, measuring add. data amounts required for second-stage training, where $1$ means the amount of data equals the training set size.
Furthermore, we examine the attention precision to evaluate whether model attention is corrected. All quantification experiments are based on five independent runs with different random seeds.

\noindent More experimental results with other architecture (ViT) and datasets (MetaShift\cite{liang2022metashift} and FMoW\cite{fmow2018}) are provided in Appx.~A.8.

\begin{table*}[htbp]
\centering
\caption{Worst-group and average accuracy ($\%$) comparison on binary classification tasks, highlighting \textbf{best} and \underline{second-best} performances. $\blacklozenge$ and $\Diamond$ indicate methods requiring spuriousness concepts and attention labels, respectively. 
} 
\label{tab:binary_results}
\resizebox{1\columnwidth}{!}{%
\begin{tabular}{p{1.8cm}<{\centering}p{2.5cm}<{\centering}p{2.8cm}<{\centering}p{1.5cm}<{\centering}p{1.5cm}<{\centering}p{1.5cm}<{\centering}p{1.5cm}<{\centering}p{1.5cm}<{\centering}p{1.5cm}<{\centering}}
\toprule[0.8pt]
                         &                              &                              & \multicolumn{2}{c}{Waterbirds} & \multicolumn{2}{c}{CelebA} & \multicolumn{2}{c}{ISIC} \\ \cmidrule(r){4-5} \cmidrule(r){6-7} \cmidrule(r){8-9}
\multirow{-2}{*}{Method} & \multirow{-2}{*}{Spuriousness label} & \multirow{-2}{*}{{\small Add. training cost}} & Worst           & Avg          & Worst         & Avg        & Worst        & Avg       \\ \hline
ERM                      &   $\usym{2717}$                        &  0$\times$                           &  $62.6_{\pm 0.3}$               & $97.3_{\pm 1.0}$              & $47.7_{\pm 2.1}$              &  $94.9_{\pm 0.3}$          &    $56.7_{\pm 2.2}$          & $79.0_{\pm 1.3}$          \\
EIIL                     & $\usym{2717}$                             &  0$\times$                            &   $83.5_{\pm 2.8}$               & $94.2_{\pm 1.3}$             &  $81.7_{\pm 0.8}$             & $85.7_{\pm 0.1}$           &  $72.3_{\pm 3.9}$            & $86.4_{\pm 1.7}$          \\
PGI                      & $\usym{2717}$                             &  0$\times$                            &  $79.5_{\pm 1.9}$               & $95.5_{\pm 0.8}$             & $85.3_{\pm 0.3}$              & $87.3_{\pm 0.1}$           & $67.4_{\pm 1.9}$              & $85.7_{\pm 1.1}$           \\
GEORGE                   & $\usym{2717}$                             &  1$\times$                           & $76.2_{\pm 2.0}$                & $95.7_{\pm 0.5}$             & $54.9_{\pm 1.9}$              & $94.6_{\pm 0.2}$           &  $58.8_{\pm 2.1}$             & $83.2_{\pm 1.4}$            \\
LfF                      & $\usym{2717}$                             &  1$\times$                           & $77.3_{\pm 2.3}$                & $91.4_{\pm 1.7}$             & $76.8_{\pm 1.2}$              &  $84.5_{\pm 0.7}$          & $72.1_{\pm 2.1}$             & $85.4_{\pm 0.5}$           \\
CIM                      & $\usym{2717}$                             & 1$\times$                             &  $77.8_{\pm 1.5}$              & $94.7_{\pm 1.1}$              &  $83.8_{\pm 1.0}$             & $90.5_{\pm 0.6}$           &  $71.7_{\pm 1.6}$            & $85.9_{\pm 1.1}$           \\
JTT                      & $\usym{2717}$                             & 5$\times$-6$\times$                            & $83.1_{\pm 3.5}$                & $90.6_{\pm 0.3}$              & $81.5_{\pm 1.7}$              &  $88.1_{\pm 0.3}$          & $28.1_{\pm 4.3}$              &  $86.2_{\pm 0.3}$         \\
CNC                      & $\usym{2717}$                             & 2$\times$-12$\times$                             &  $88.5_{\pm 0.3}$                 &  $90.9_{\pm 0.1}$             & $88.8_{\pm 0.9}$              &  $89.9_{\pm 0.5}$          & $68.2_{\pm 0.2}$             & $84.6_{\pm 0.2}$          \\ \bottomrule[0.1pt]
DISC                      & $\usym{2717}$ $\blacklozenge$                           & 1$\times$                            & $88.7_{\pm 0.7}$                   &  $93.8_{\pm 0.7}$            &  $85.7_{\pm 1.1}$             & $89.9_{\pm 0.7}$          &  $73.4_{\pm 0.4}$           &  $84.7_{\pm 0.4}$          \\

\rowcolor[HTML]{EFEFEF} 
\slimTr                  & $\usym{2717}$ $\Diamond$                            & 0.1$\times$-0.2$\times$                            &  $89.1_{\pm 0.6}$                &  $91.7_{\pm 0.5}$            &  \underline{$89.4_{\pm 0.3}$}             & $90.1_{\pm 0.2}$           & \bm{$79.1_{\pm 0.5}$}             &  $88.5_{\pm 0.4}$         \\ 
\rowcolor[HTML]{EFEFEF} 
\slimVal                 & $\usym{2717}$  $\Diamond$                             &0.05$\times$-0.3$\times$                                  & \bm{$91.0_{\pm 0.4}$}                &$93.8_{\pm 0.4}$              &  \underline{$89.4_{\pm 0.5}$}               & $90.7_{\pm 0.2}$             & \underline{$74.0_{\pm 0.2}$}            &  $85.4_{\pm 0.2}$         \\
\bottomrule[0.1pt]
SSA                      & validation                             & 1.5$\times$-5$\times$                              &   $89.0_{\pm 0.6}$                & $92.2_{\pm 0.9}$              & \bm{$89.8_{\pm 1.3}$}              & $92.8_{\pm 0.1}$           &   $61.7_{\pm 0.7}$           & $77.7_{\pm 1.0}$          \\
DFR                      & validation                             & 0.05$\times$-0.3$\times$                             &  $89.6_{\pm 0.7}$                & $92.1_{\pm 0.2}$             & $87.3_{\pm 1.0}$              & $90.2_{\pm 0.8}$           &  $71.9_{\pm 0.7}$             & $84.3_{\pm 0.3}$          \\
 \bottomrule[0.1pt]
GDRO                     & training                             &0$\times$                             &  \underline{$89.9_{\pm 0.6}$}               &  $92.6_{\pm 0.1}$            &    $88.9_{\pm 1.3}$             & $93.9_{\pm 0.1}$             & $66.8_{\pm 3.5}$              & $80.5_{\pm 1.9}$          \\ 
PDE                     & training                             &0$\times$                                & $85.9_{\pm 2.7}$                  &  $90.5_{\pm 1.8}$            &  $85.0_{\pm 4.0}$               &  $92.3_{\pm 0.8}$              & $66.4_{\pm 2.3}$             & $82.2_{\pm 1.3}$         \\ 
\bottomrule[0.8pt]
\end{tabular}%
}
\end{table*}
\begin{table*}[htbp]
\caption{Worst-group and average accuracy ($\%$) comparison on multi-classification tasks, highlighting \textbf{best} and \underline{second-best} results. $\Diamond$ indicates methods requiring attention labels. - represents inapplicable methods because such methods require spuriousness labels while the dataset does not provide them.}
\centering
\label{tab:multi_results}
\resizebox{0.8\columnwidth}{!}{%
\begin{tabular}{p{1.5cm}<{\centering}p{2.5cm}<{\centering}p{3.5cm}<{\centering}p{1.5cm}<{\centering}p{1.5cm}<{\centering}p{1.5cm}<{\centering}p{1.5cm}<{\centering}}
\toprule[0.8pt]
                         &                              &                              & \multicolumn{2}{c}{NICO} & \multicolumn{2}{c}{ImageNet9} \\ \cmidrule(r){4-5}  \cmidrule(r){6-7}
\multirow{-2}{*}{Method} & \multirow{-2}{*}{Spuriousness label} & \multirow{-2}{*}{{\small Add. training cost}} & Worst        & Avg       & Worst          & Avg          \\ \hline
ERM                      &  $\usym{2717}$                             & 0$\times$                              &  $11.9_{\pm 0.2}$             &  $70.9_{\pm 1.2}$          &   $55.3_{\pm 2.0}$              &   $75.6_{\pm 1.1}$            \\
CNC                      & $\usym{2717}$                              & 5$\times$-12$\times$                              & $72.6_{\pm 1.7}$             & $89.5_{\pm 0.6}$          &  $58.9_{\pm 0.6}$              &  $74.3_{\pm 0.2}$             \\\bottomrule[0.1pt]
\rowcolor[HTML]{EFEFEF} 
\slimTr                & $\usym{2717}$ $\Diamond$                             & 0.1$\times$-0.2$\times$                             & \bm{$78.0_{\pm 0.8}$}            &  $92.1_{\pm 0.1}$         &   \bm{$68.1_{\pm 0.3}$}             & $74.5_{\pm 0.2}$             \\
\rowcolor[HTML]{EFEFEF} 
\slimVal                 &  $\usym{2717}$ $\Diamond$                          & 0.1$\times$-0.2$\times$                             &  \underline{$73.8_{\pm 0.4}$}            &  $90.7_{\pm 0.2}$          &  \underline{$64.6_{\pm 0.4}$}               &  $74.2_{\pm 0.3}$             \\ \bottomrule[0.1pt]
DFR                      & validation                             & 0.4$\times$                           & $70.9_{\pm 2.1}$              & $85.8_{\pm 0.6}$          & -               & -             \\  \hline

GDRO                     &    training                          &0$\times$                              &  $73.3_{\pm 1.0}$            & $90.3_{\pm 0.5}$           &  -              &  -            \\ \bottomrule[0.8pt]
\end{tabular}%
}
\end{table*}

\begin{table*}[tbhp]
\centering
\caption{\slim's primary cost, including the data amount required for attention annotation \textit{(Att.)}, and the constructed data size for model training \textit{(Training)}.}
\label{tab:slim_primary_cost}
\resizebox{1\columnwidth}{!}{%
\begin{tabular}{p{1.5cm}<{\centering}p{1.8cm}<{\centering}p{1.8cm}<{\centering}p{1.8cm}<{\centering}p{1.8cm}<{\centering}p{1.8cm}<{\centering}p{1.8cm}<{\centering}p{1.8cm}<{\centering}p{1.8cm}<{\centering}p{1.8cm}<{\centering}p{1.8cm}<{\centering}}
\toprule[0.8pt]
\multirow{2}{*}{Method} & \multicolumn{2}{c}{Waterbird} & \multicolumn{2}{c}{CelebA} & \multicolumn{2}{c}{ISIC} & \multicolumn{2}{c}{NICO} & \multicolumn{2}{c}{ImageNet 9}  \\ \cmidrule(r){2-3} \cmidrule(r){4-5} \cmidrule(r){6-7} \cmidrule(r){8-9}\cmidrule(r){10-11}
                        & Att.    & Training    & Att.   & Training  & Att.   & Training  & Att.   & Training & Att.   & Training \\ \hline
\slimTr               &  $120$ ($2.5\%$ )            & $0.1\times$             & $200$ ($0.12\%$)             & $0.1\times$             & $120$ ($0.7\%$)            &  $0.2\times$              & $40$ ($1.8\%$)             & $0.2\times$             & $180$ ($0.4\%$)            &  $0.1\times$ \\
\slimVal                  & $40$ ($3\%$ )              & $0.1\times$                &  $70$ ($0.35\%$)            &  $0.05\times$            & $40$ ($2\%$)            & $0.3\times$             & $20$ ($2.7\%$)             & $0.2\times$             & $50$ ($1\%$)            &  $0.1\times$ \\ \bottomrule[0.8pt]
\end{tabular}%
}
\end{table*}

\subsection{Spuriousness Mitigation Performance (Q1)}
\label{subsec:overall_perform}

In this section, we present the performance evaluation by comparing different methods across multiple binary and multiclass classification datasets mentioned in Sec.~\ref{subsec:experimental_setup}. Key findings are demonstrated below.

\noindent\textbf{Superior worst-group performance.}
The metric of worst-group accuracy ($Acc_w$) is pivotal for spuriousness mitigation methods (refer to Sec.~\ref{sec:problem}), and the results in Tables~\ref{tab:binary_results} and~\ref{tab:multi_results} showcase~\slim's superior capability in this regard. Across various classification tasks,~\slim~consistantly outperforms competing methods in achieving the highest $Acc_w$ while maintaining high average accuracy. Specifically, in binary classification scenarios (Table~\ref{tab:binary_results}),~\slim~leads in performance for Waterbirds and ISIC and secures a close second in CelebA. In multi-classification tasks (Table~\ref{tab:multi_results}), \slimTr~and \slimVal~rank as the top performers across the board, underscoring the effectiveness of our approach in spuriousness mitigation.

\noindent\textbf{Cost-efficient spuriousness mitigation.}
Methods in Tables~\ref{tab:binary_results} and~\ref{tab:multi_results} are divided into four groups from top to bottom, as per the categorization established in Sec.~\ref{subsec:experimental_setup} according to the computational cost. The \textit{``Add. training cost''} columns denote the additional training cost for mitigating spuriousness apart from the initial training, such as additional learning phases ({\small CNC}, {\small SSA}) or data augmentation ({\small JTT}).
Results in the table substantiate that methods with higher costs (higher \textit{Add. training costs} or more spuriousness labels) tend to outperform others, aligning with prior research findings. Nevertheless, among the well-performed methods, it is worth mentioning that~\slim~is more cost-efficient: \slimTr~used $0.1$-$0.2$ proportion of training sets and \slimVal~used $0.05$-$0.3$ proportion of validation sets for additional training, which is the same as {\small DFR} and much smaller than others. Moreover, \slim~requires labels for attention correctness, which is more cost-efficient to obtain than spuriousness labels (refer to Sec.~\ref{subsec:att_eval_performance} for more details).
\begin{table*}[bth]
\caption{Attention accuracy evaluation. (Left) {\small GradCAM} visualization of model attention. (Right) Worst group and average attention accuracy of selected methods.}
\centering
\begin{minipage}{0.66\textwidth}
    \centering
    \includegraphics[width=0.95\linewidth]{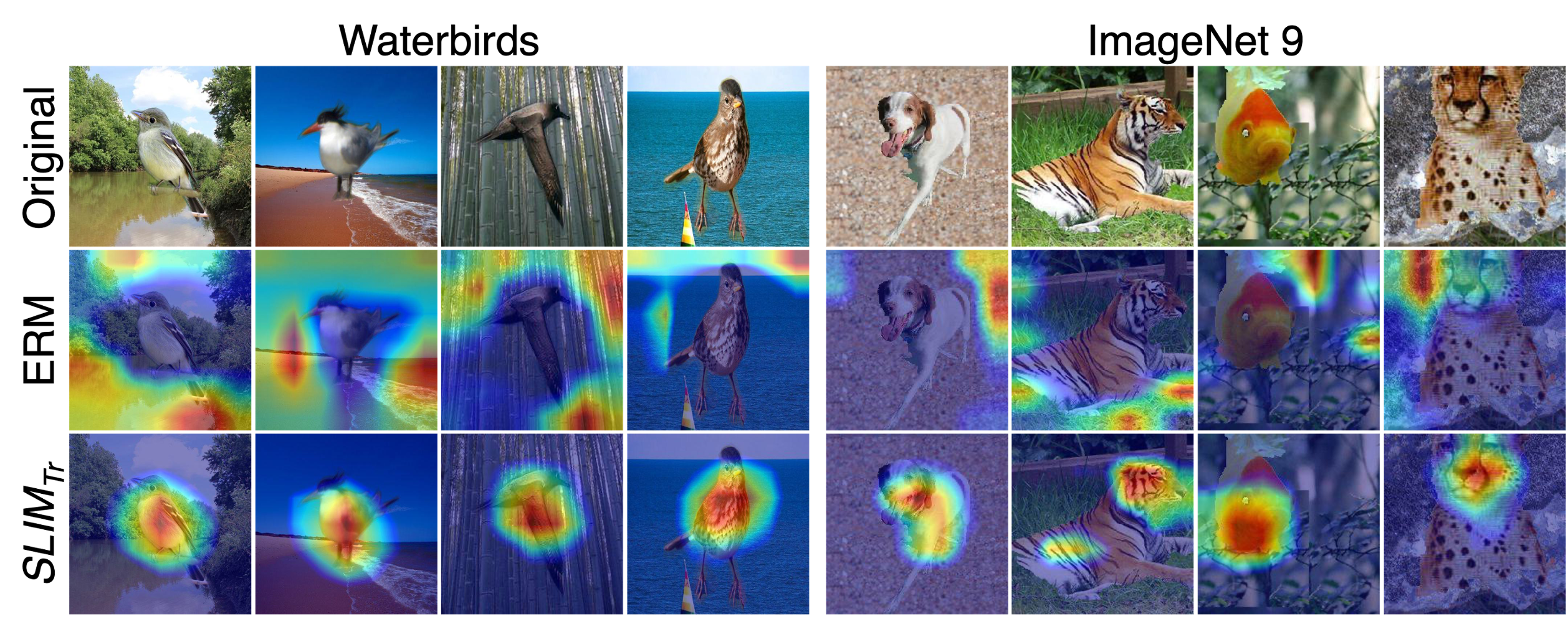}
\end{minipage}\hfill
\begin{minipage}{0.3\textwidth}
    \centering
    \resizebox{0.75\columnwidth}{!}{%
    \begin{tabular}{p{1.7cm}<{\centering}p{1.4cm}<{\centering}p{1.4cm}<{\centering}}
\toprule[0.8pt]
\multirow{2}{*}{Waterbirds} & \multicolumn{2}{c}{AIoU} \\ \cmidrule(r){2-3} 
                            & Worst       & Avg        \\ \hline
ERM                         & $.25_{\pm .05}$       & $.47_{\pm .03}$      \\
CNC                         & $.48_{\pm .05}$       & $.49_{\pm .04}$    \\
GDRO                        & $.49_{\pm .03}$       & $.52_{\pm .03}$      \\
PDE                         & $.50_{\pm .02}$       & $.52_{\pm .03}$       \\
\slimTr                        & $.53_{\pm .02}$       & $.61_{\pm .02}$       \\ \hline\hline
\multirow{2}{*}{ImageNet 9} & \multicolumn{2}{c}{AIoU} \\ \cmidrule(r){2-3}
                            & Worst       & Avg        \\ \hline
ERM                         & $.33_{\pm .07}$        & $.55_{\pm .03}$      \\
CNC                         & $.44_{\pm .03}$        &  $.61_{\pm .03}$      \\
\slimTr                        & $.58_{\pm .04}$        & $.68_{\pm .03}$       \\ \bottomrule[0.8pt]
\end{tabular}%
            }\label{tab:att_compare}
    \end{minipage}
\end{table*}
Table~\ref{tab:slim_primary_cost} presents~\slim's primary cost, including the training data amount for mitigating spuriousness, which is the detailed version of \textit{``Add. training cost''} in Tables~\ref{tab:binary_results} and~\ref{tab:multi_results}, and the data amount requiring attention labels \textit{(Att.)}. Notably, our approach only requires no more than $3\%$ of the corresponding data split (train or validation) for attention-label annotation, further proving the cost-efficiency of~\slim.

\noindent\textbf{Enhanced attention accuracy.}
Beyond classification accuracy, we further evaluate the performance of spuriousness mitigation methods via attention accuracy. Specifically, for models resulting from each method, we randomly sample data instances to visualize model attention, and measure attention accuracy scores in both worst-group and dataset levels, consistent with our evaluation of classification accuracy. Align with the related work\cite{10.5555/3618408.3620051}, we employ the Adjusted Intersection-over-Union (AIoU) score as our attention accuracy measurement, which is an adjusted IoU score with a threshold, aiming at measuring the overlap between model's highest attributed region and the ground-truth bounding box. Details related to AIoU are given in Appx.~A.7. Table~\ref{tab:att_compare} presents the evaluation results. The left {\small GradCAM} examples showcase \slim's capability of correcting model's wrong attention, and the right table compares \slim~with SOTA methods, where we can find our method secures the best performance in terms of both worst-group and average attention accuracy.

\subsection{Attention Labeling Performance (Q2)}\label{subsec:att_eval_performance}
To compare the processes of annotating spuriousness versus attention correctness, we employed crowdsourcing tasks using the Waterbirds and NICO datasets, as both of which come with ground-truth spuriousness labels. For this study, we selected a random set of 120 images from each dataset and established two separate tasks: one for spuriousness labeling and one for attention correctness labeling. For each task, we recruited 60 workers independently from the Prolific platform to prevent learning biases from cross-task participation. We meticulously recorded the time taken to label each image and reviewed all submissions to exclude any outliers with unusually long or short annotation times. Ensuring ethical research standards, our study avoided collecting personally identifiable information and excluded any potentially offensive content. Detailed instructions provided to the crowdsourcing participants can be found in Appx.~A.6.

\noindent\textbf{Evaluation Consistency.} 
Table~\ref{tab:human_annotation_eval} presents metrics for this study. While the annotation time for Waterbirds remains similar across tasks, attention correctness for NICO is annotated more quickly than the other. This difference may stem
\begin{minipage}{\textwidth}
\begin{minipage}[htbp]{0.4\textwidth}
\centering
\makeatletter\def\@captype{table}\makeatother\caption{Comparison of crowdsourced labeling tasks: spuriousness vs. attention correctness.}
\resizebox{1\columnwidth}{!}{
\begin{tabular}{p{2.8cm}<{\centering}p{1.8cm}<{\centering}p{2cm}<{\centering}}
\toprule[0.8pt]
 Waterbirds & Avg. Time & Avg. Consis. \\ \hline
Spuriousness & $4.9_{\pm 2.3}$ sec         & $81.9_{\pm 14.8}\%$                 \\
Att. &  $4.5_{\pm 1.4}$ sec         &  $94.6_{\pm 10.5}\%$                \\ \hline\hline
 NICO & Avg. Time & Avg. Consis. \\ \hline
Spuriousness & $7.9_{\pm 3.8}$ sec         & $70.1_{\pm 19.1}\%$                 \\
Att. &  $5.3_{\pm 1.6}$ sec         &  $95.1_{\pm 6.1}\%$\\ \bottomrule[0.8pt]
\end{tabular}
 }\label{tab:human_annotation_eval}
  \end{minipage}
\begin{minipage}[htbp]{0.54\textwidth}
   \centering
   \includegraphics[height=0.31\textwidth]{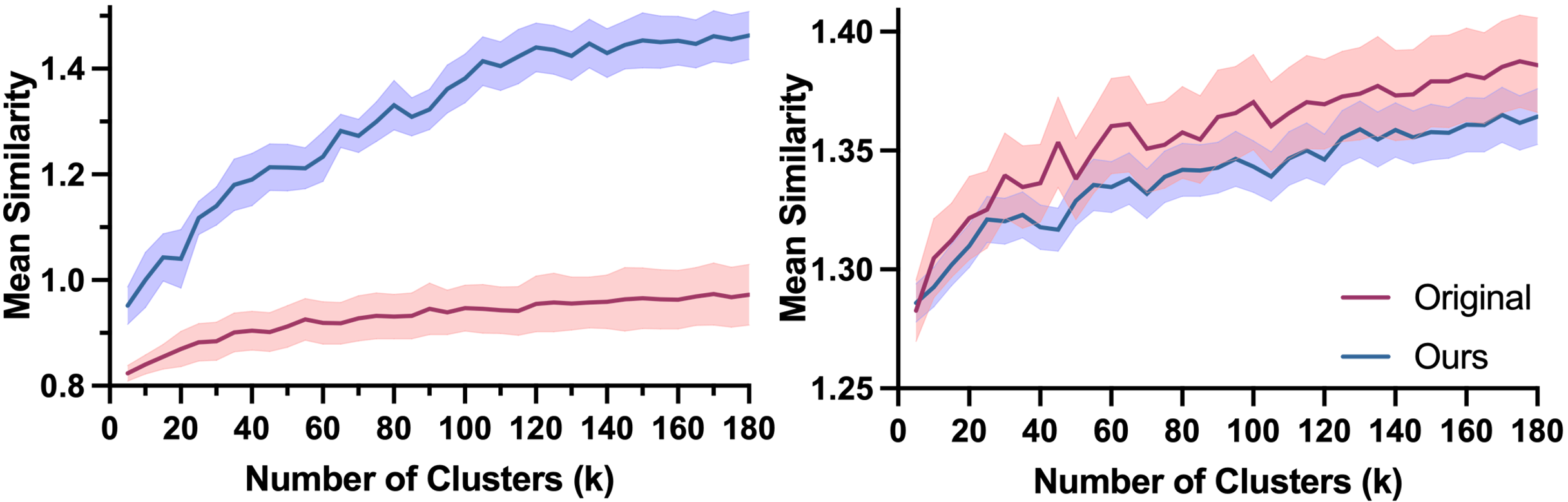}
   \makeatletter\def\@captype{figure}\makeatother\caption{Comparison of neighbor consistency over {\color{BrickRed} original feature space} and {\color{MidnightBlue} our attention space} on Waterbirds, w.r.t. model attention (left) and data features (right).}
   \label{fig:vecsim}
  \end{minipage}
\end{minipage}
from NICO's broader array of spuriousness choices versus the binary attention annotations.
To measure the annotation consistency—whether a single receives divergent labels—we use the percentage of the most frequent label as a consensus measure. A higher percentage suggests greater consistency. Our findings indicate that attention correctness annotations exhibit a notably higher consistency, suggesting that it is less ambiguous to determine correct attention than to identify spurious features.

\noindent\textbf{Attention Consistency.} \slim~requires spreading labels in a space where attention patterns are consistently similar in nearby areas. To evaluate its feasibility, we compared how similar neighbors are in the original feature space versus the attention space created by \slim. We used k-means clustering with varying numbers of clusters to measure the intra-cluster similarity for both the model's attention matrices and the feature vectors. A higher similarity score means the space has more consistent patterns, which is essential for the accurate spreading of labels. As Fig.~\ref{fig:vecsim} shows, our attention space significantly improves consistency in the model's attention, while also maintaining a good level of data feature consistency. Increasing the number of clusters generally increases consistency, but there is a point where adding more clusters does not lead to further gains, as shown by the curve flattening at $k$=$120$ in the left plot.

In summary, we have demonstrated that the annotation module of \slim is less time-consuming and yields more consistent labeling feedback. Additionally, the attention space engineered by \slim~establishes a solid foundation for propagating attention labels, considering the consistent model attention observed among neighboring instances.

\subsection{Ablation Study (Q3)}\label{subsec:ablation_study}
We investigate two critical factors of \slim, which are detailed below. Other factors such as the dimension of the attention space are discussed in Appx.~A.8.

\noindent\textbf{Influence of data amount for attention annotations.}
We change the data amount for attention annotations to investigate their effect on performance, keeping other variables constant. Table~\ref{tab:eva_varying} illustrates that both the worst-group and average accuracy tend to rise with more annotations, while such incremental benefit stabilizes after reaching a certain point, helping us identify an optimal number of annotations that balance performance with annotation cost.

\begin{table*}[ht]
\centering
\caption{
Performance variation with different attention annotation amounts in \slim. Marks in table: \textbf{best} and \underline{second best} performance; the \colorbox{LB}{amounts} used in Table~\ref{tab:binary_results}.
}
\label{tab:eva_varying}
\resizebox{0.86\columnwidth}{!}{%
\begin{tabular}{p{1.3cm}<{\centering}p{1.5cm}<{\centering}p{1.3cm}<{\centering}p{1.3cm}<{\centering}p{1.3cm}<{\centering}p{1.3cm}<{\centering}p{1.3cm}<{\centering}p{1.3cm}<{\centering}p{1.3cm}<{\centering}p{1.3cm}<{\centering}}
\toprule[0.8pt]
\multirow{2}{*}{Method}   & \multicolumn{3}{c}{Waterbirds} & \multicolumn{3}{c}{CelebA} & \multicolumn{3}{c}{ISIC} \\  \cmidrule(r){2-4}  \cmidrule(r){5-7}  \cmidrule(r){8-10} 
                          & Amount     & Worst     & Avg    & Amount    & Worst   & Avg   & Amount   & Worst   & Avg  \\ \hline
\multirow{5}{*}{\slimTr} & 60         & $78.2_{\pm 0.5}$           & $86.0_{\pm 0.3}$       & 100        &$86.4_{\pm 0.2}$          & $90.2_{\pm 0.1}$      & 60       &$75.3_{\pm 0.2}$           &$86.6_{\pm 0.9}$       \\
                          & 90         &$85.9_{\pm 0.9}$              &  $89.3_{\pm 0.3}$        & 150       & $87.8_{\pm 0.4}$        &$91.1_{\pm 0.7}$     & 90      &$74.4_{\pm 0.4}$           &$87.4_{\pm 0.7}$     \\
                          & \cellcolor[HTML]{ECF4FF}120         & $89.1_{\pm 0.6}$           & $91.7_{\pm 0.5}$       & \cellcolor[HTML]{ECF4FF}200       &$89.4_{\pm 0.4}$            &$90.1_{\pm 0.2}$     & \cellcolor[HTML]{ECF4FF}120      & $79.1_{\pm 0.5}$         &$88.5_{\pm 0.4}$      \\
                          & 150        & \bm{$89.4_{\pm 0.6}$}           & $91.7_{\pm 0.2}$        & 250       &\underline{$89.4_{\pm 0.2}$}         & $90.9_{\pm 0.1}$      & 150      &\bm{$79.9_{\pm 0.8}$}         &$88.8_{\pm 0.7}$      \\
                          & 180         &   \underline{$89.3_{\pm 0.3}$}          &  $91.0_{\pm 1.5}$       & 300       & \bm{$89.5_{\pm 0.1}$}        &$91.2_{\pm 0.2}$        & 180      &  \underline{$79.8_{\pm 0.3}$}       &  $88.6_{\pm 0.3}$    \\ \bottomrule[0.1pt]
\multirow{5}{*}{\slimVal} & 10         &$78.9_{\pm 8.7}$           & $86.5_{\pm 5.1}$       & 40        & $84.7_{\pm 0.4}$         & $87.4_{\pm 0.3}$       & 10       & $70.5_{\pm 0.5}$        & $83.3_{\pm 1.2}$     \\
                          & 25        & $89.2_{\pm 0.6}$          & $93.4_{\pm 1.0}$         & 55       & $87.6_{\pm 0.3}$        &$90.4_{\pm 0.6}$        & 25      & $72.7_{\pm 0.1}$        &$85.2_{\pm 0.8}$     \\
                          & \cellcolor[HTML]{ECF4FF}40        & \underline{$91.0_{\pm 0.4}$}          & $93.8_{\pm 0.4}$       & \cellcolor[HTML]{ECF4FF}70       &$89.4_{\pm 0.4}$         & $90.7_{\pm 0.2}$      & \cellcolor[HTML]{ECF4FF}40      & \underline{$74.0_{\pm 0.2}$}        & $85.4_{\pm 0.2}$      \\
                          & 55        & \bm{$91.5_{\pm 0.3}$}          &$93.8_{\pm 0.1}$        & 85       & \bm{$89.7_{\pm 0.1}$}         & $90.5_{\pm 0.4}$      & 55      & \underline{$74.0_{\pm 0.2}$}        &$85.9_{\pm 0.5}$      \\
                          & 70        & $90.7_{\pm 0.3}$          & $94.0_{\pm 0.3}$        & 100       & \underline{$89.7_{\pm 0.2}$}        & $90.5_{\pm 0.2}$       & 70      & \bm{$74.1_{\pm 0.2}$}         & $85.8_{\pm 0.2}$     \\ \bottomrule[0.8pt]
\end{tabular}%
}
\end{table*}

\begin{table*}[ht]
\centering
\caption{
Performance variation with different training sizes. \slimTr~and \slimVal~are trained on different proportions (Prop) of training sets and validation sets, respectively.
}
\label{tab:train_varying}
\resizebox{0.86\columnwidth}{!}{%
\begin{tabular}{p{1.3cm}<{\centering}p{1.3cm}<{\centering}p{1.3cm}<{\centering}p{1.3cm}<{\centering}p{1.3cm}<{\centering}p{1.3cm}<{\centering}p{1.3cm}<{\centering}p{1.3cm}<{\centering}p{1.3cm}<{\centering}p{1.3cm}<{\centering}}
\toprule[0.8pt]
\multirow{2}{*}{Method}   & \multicolumn{3}{c}{Waterbirds} & \multicolumn{3}{c}{CelebA} & \multicolumn{3}{c}{ISIC} \\  \cmidrule(r){2-4}  \cmidrule(r){5-7}  \cmidrule(r){8-10} 
                          & Prop$\%$     & Worst     & Avg    & Prop$\%$    & Worst   & Avg   & Prop$\%$   & Worst   & Avg  \\ \hline
\multirow{4}{*}{\slimTr} & 5         & $82.4_{\pm 1.0}$          & $83.6_{\pm 1.3}$        & 5        & $88.9_{\pm 0.1}$        & $90.0_{\pm 0.1}$       & 15       &$77.6_{\pm 2.1}$         &$87.1_{\pm 1.2}$       \\
                          & \cellcolor[HTML]{ECF4FF}10        & \bm{$89.1_{\pm 0.6}$}           & $91.7_{\pm 0.5}$       & \cellcolor[HTML]{ECF4FF}10       & \bm{$89.4_{\pm 0.3}$}       & $90.1_{\pm 0.2}$      & \cellcolor[HTML]{ECF4FF}20      & \bm{$79.1_{\pm 0.5}$}         & $88.5_{\pm 0.4}$     \\
                          & 12       &  $82.3_{\pm 1.3}$          & $87.8_{\pm 0.3}$       & 15       &  $89.4_{\pm 0.4}$      & $91.0_{\pm 0.4}$      & 25     & $77.4_{\pm 1.5}$        & $88.2_{\pm 1.2}$     \\
                          & 15        & $83.5_{\pm 2.7}$           & $89.6_{\pm 0.5}$       & 20       & $88.9_{\pm 0.1}$        &  $90.8_{\pm 0.3}$     & 30     & $78.3_{\pm 1.5}$        & $88.3_{\pm 0.9}$     \\ \hline
\multirow{4}{*}{\slimVal} & 5         & $84.2_{\pm 3.6}$           & $89.2_{\pm 2.6}$        & 3        &$85.7_{\pm 1.3}$          & $88.5_{\pm 1.1}$      & 25       & $68.2_{\pm 2.1}$        & $81.9_{\pm 0.8}$     \\
                          & \cellcolor[HTML]{ECF4FF}10        & \bm{$91.0_{\pm 0.4}$}           & $93.8_{\pm 0.4}$        & \cellcolor[HTML]{ECF4FF}5       &\bm{$89.4_{\pm 0.4}$}          & $90.7_{\pm 0.2}$       & \cellcolor[HTML]{ECF4FF}30     & \bm{$74.0_{\pm 0.2}$}        &  $85.4_{\pm 0.2}$     \\
                          & 12       &  $87.6_{\pm 0.6}$         & $93.9_{\pm 1.3}$       & 10       & $88.3_{\pm 1.4}$        & $90.3_{\pm 0.5}$      & 35      & $72.5_{\pm 0.5}$        & $85.3_{\pm 0.1}$     \\
                          & 15        &$86.8_{\pm 2.0}$           & $94.6_{\pm 0.5}$       & 15       &$87.2_{\pm 0.4}$         &$90.3_{\pm 0.4}$        & 40      & $70.9_{\pm 1.2}$         & $85.4_{\pm 3.5}$     \\ \bottomrule[0.8pt]
\end{tabular}%
}
\end{table*}

\noindent\textbf{Influence of training set size.}
This size refers to the size of the data subset constructed in \slim's final phase for training a robust model. By only varying this parameter (Table~\ref{tab:train_varying}), we find that the optimal size is inherently linked to the size of our constructed subgroups, $|g_{\hat{c}_i, e_i}|$. An optimal size is reached when data from various groups are sufficiently sampled without redundant oversampling.
These findings guide the hyperparameter selection in this phase, ensuring an effective cost allocation.

\section{Conclusion}\label{sec:conclusion}
In this paper, we introduce \slim, a novel framework for spuriousness mitigating in deep learning. With minimal human annotation effort, \slim~provides a data construction pipeline to create a feature-balanced subset for training a robust model. Extensive evaluation demonstrates the superior cost-effectiveness of \slim, which matches or outperforms state-of-the-art methods with far less costs. The success of \slim~underscores the potential of data quality and efficient human supervision in the development of robust models. We believe that \slim~will inspire further research in this direction, ultimately leading to more cost-effective machine learning systems for reliable AI.

\noindent\textbf{Limitations.} A limitation of our method is its reliance on attention-based spuriousness detection. Despite its effectiveness in handling spurious features that can be represented by a certain image region, it overlooks some types of spurious features, such as color or lighting. Such features are hard to be disentangled by attention-based model attributions. In the future, we plan to study how to mitigate other formats of spurious features.

\noindent\textbf{Acknowledgements}
This research is supported in part by Bosch Research and the National Institute of Health with grants P41-EB032840 and R01CA270454.
H.-T. Lin is partially supported by the National Taiwan University Center for Data Intelligence via NTU-112L900901 and the Ministry of Science and Technology in Taiwan via MOST 112-2628-E-002-030.

%
%
\bibliographystyle{splncs04}
\bibliography{main}

\begin{thebibliography}{10}
\providecommand{\url}[1]{\texttt{#1}}
\providecommand{\urlprefix}{URL }
\providecommand{\doi}[1]{https://doi.org/#1}

\bibitem{ahmed2020systematic}
Ahmed, F., Bengio, Y., Van~Seijen, H., Courville, A.: Systematic generalisation with group invariant predictions. In: International Conference on Learning Representations (2020)

\bibitem{bontempelli2023conceptlevel}
Bontempelli, A., Teso, S., Tentori, K., Giunchiglia, F., Passerini, A.: Concept-level debugging of part-prototype networks. In: The Eleventh International Conference on Learning Representations (2023), \url{https://openreview.net/forum?id=oiwXWPDTyNk}

\bibitem{cao2022benign}
Cao, Y., Chen, Z., Belkin, M., Gu, Q.: Benign overfitting in two-layer convolutional neural networks. Advances in neural information processing systems  \textbf{35},  25237--25250 (2022)

\bibitem{chen2022towards}
Chen, Z., Deng, Y., Wu, Y., Gu, Q., Li, Y.: Towards understanding the mixture-of-experts layer in deep learning. Advances in neural information processing systems  \textbf{35},  23049--23062 (2022)

\bibitem{fmow2018}
Christie, G., Fendley, N., Wilson, J., Mukherjee, R.: Functional map of the world. In: CVPR (2018)

\bibitem{8363547}
Codella, N.C.F., Gutman, D., Celebi, M.E., Helba, B., Marchetti, M.A., Dusza, S.W., Kalloo, A., Liopyris, K., Mishra, N., Kittler, H., Halpern, A.: Skin lesion analysis toward melanoma detection: A challenge at the 2017 international symposium on biomedical imaging (isbi), hosted by the international skin imaging collaboration (isic). In: 2018 IEEE 15th International Symposium on Biomedical Imaging (ISBI 2018). pp. 168--172 (2018). \doi{10.1109/ISBI.2018.8363547}

\bibitem{creager2021environment}
Creager, E., Jacobsen, J.H., Zemel, R.: Environment inference for invariant learning. In: International Conference on Machine Learning. pp. 2189--2200. PMLR (2021)

\bibitem{deng2023robust}
Deng, Y., Yang, Y., Mirzasoleiman, B., Gu, Q.: Robust learning with progressive data expansion against spurious correlation. In: Thirty-seventh Conference on Neural Information Processing Systems (2023), \url{https://openreview.net/forum?id=9QEVJ9qm46}

\bibitem{dosovitskiy2020image}
Dosovitskiy, A., Beyer, L., Kolesnikov, A., Weissenborn, D., Zhai, X., Unterthiner, T., Dehghani, M., Minderer, M., Heigold, G., Gelly, S., et~al.: An image is worth 16x16 words: Transformers for image recognition at scale. arXiv preprint arXiv:2010.11929  (2020)

\bibitem{10.1145/3555590}
Gao, Y., Sun, T.S., Zhao, L., Hong, S.R.: Aligning eyes between humans and deep neural network through interactive attention alignment. Proc. ACM Hum.-Comput. Interact.  \textbf{6}(CSCW2) (nov 2022). \doi{10.1145/3555590}, \url{https://doi.org/10.1145/3555590}

\bibitem{gao2022aligning}
Gao, Y., Sun, T.S., Zhao, L., Hong, S.R.: Aligning eyes between humans and deep neural network through interactive attention alignment. Proceedings of the ACM on Human-Computer Interaction  \textbf{6}(CSCW2),  1--28 (2022)

\bibitem{ghosal2024vision}
Ghosal, S.S., et~al.: Are vision transformers robust to spurious correlations? IJCV  \textbf{132}(3),  689--709 (2024)

\bibitem{he2009learning}
He, H., Garcia, E.A.: Learning from imbalanced data. IEEE Transactions on knowledge and data engineering  \textbf{21}(9),  1263--1284 (2009)

\bibitem{jelassi2022towards}
Jelassi, S., Li, Y.: Towards understanding how momentum improves generalization in deep learning. In: International Conference on Machine Learning. pp. 9965--10040. PMLR (2022)

\bibitem{joshi2023towards}
Joshi, S., Yang, Y., Xue, Y., Yang, W., Mirzasoleiman, B.: Towards mitigating spurious correlations in the wild: A benchmark \& a more realistic dataset. arXiv preprint arXiv:2306.11957  (2023)

\bibitem{kirichenko2022last}
Kirichenko, P., Izmailov, P., Wilson, A.G.: Last layer re-training is sufficient for robustness to spurious correlations. arXiv preprint arXiv:2204.02937  (2022)

\bibitem{leetowards}
Lee, S., Payani, A., Chau, D.H.P.: Towards mitigating spurious correlations in image classifiers with simple yes-no feedback. In: AI and HCI Workshop at the 40th International Conference on Machine Learning (2023)

\bibitem{liang2022metashift}
Liang, W., Zou, J.: Metashift: A dataset of datasets for evaluating contextual distribution shifts and training conflicts. In: International Conference on Learning Representations (2022), \url{https://openreview.net/forum?id=MTex8qKavoS}

\bibitem{liu2021just}
Liu, E.Z., Haghgoo, B., Chen, A.S., Raghunathan, A., Koh, P.W., Sagawa, S., Liang, P., Finn, C.: Just train twice: Improving group robustness without training group information. In: International Conference on Machine Learning. pp. 6781--6792. PMLR (2021)

\bibitem{liu2021swin}
Liu, Z., Lin, Y., Cao, Y., Hu, H., Wei, Y., Zhang, Z., Lin, S., Guo, B.: Swin transformer: Hierarchical vision transformer using shifted windows. In: Proceedings of the IEEE/CVF international conference on computer vision. pp. 10012--10022 (2021)

\bibitem{liu2015deep}
Liu, Z., Luo, P., Wang, X., Tang, X.: Deep learning face attributes in the wild. In: Proceedings of the IEEE international conference on computer vision. pp. 3730--3738 (2015)

\bibitem{mcinnes2018umap}
McInnes, L., Healy, J., Melville, J.: Umap: Uniform manifold approximation and projection for dimension reduction. arXiv preprint arXiv:1802.03426  (2018)

\bibitem{nam2020learning}
Nam, J., Cha, H., Ahn, S., Lee, J., Shin, J.: Learning from failure: De-biasing classifier from biased classifier. Advances in Neural Information Processing Systems  \textbf{33},  20673--20684 (2020)

\bibitem{nam2022spread}
Nam, J., Kim, J., Lee, J., Shin, J.: Spread spurious attribute: Improving worst-group accuracy with spurious attribute estimation. arXiv preprint arXiv:2204.02070  (2022)

\bibitem{nguyen2021the}
Nguyen, G., Kim, D., Nguyen, A.: The effectiveness of feature attribution methods and its correlation with automatic evaluation scores. In: Beygelzimer, A., Dauphin, Y., Liang, P., Vaughan, J.W. (eds.) Advances in Neural Information Processing Systems (2021), \url{https://openreview.net/forum?id=OKPS9YdZ8Va}

\bibitem{petsiuk2018rise}
Petsiuk, V., Das, A., Saenko, K.: Rise: Randomized input sampling for explanation of black-box models. arXiv preprint arXiv:1806.07421  (2018)

\bibitem{rao2023using}
Rao, S., B{\"o}hle, M., Parchami-Araghi, A., Schiele, B.: Using explanations to guide models. arXiv preprint arXiv:2303.11932  (2023)

\bibitem{10.5555/3524938.3525689}
Rieger, L., Singh, C., Murdoch, W.J., Yu, B.: Interpretations are useful: Penalizing explanations to align neural networks with prior knowledge. In: Proceedings of the 37th International Conference on Machine Learning. ICML'20, JMLR.org (2020)

\bibitem{ijcai2017p371}
Ross, A.S., Hughes, M.C., Doshi-Velez, F.: Right for the right reasons: Training differentiable models by constraining their explanations. In: Proceedings of the Twenty-Sixth International Joint Conference on Artificial Intelligence, {IJCAI-17}. pp. 2662--2670 (2017). \doi{10.24963/ijcai.2017/371}, \url{https://doi.org/10.24963/ijcai.2017/371}

\bibitem{Sagawa2020Distributionally}
Sagawa*, S., Koh*, P.W., Hashimoto, T.B., Liang, P.: Distributionally robust neural networks. In: International Conference on Learning Representations (2020), \url{https://openreview.net/forum?id=ryxGuJrFvS}

\bibitem{schramowski2020making}
Schramowski, P., et~al.: Making deep neural networks right for the right scientific reasons by interacting with their explanations. Nature  \textbf{2}(8),  476--486 (2020)

\bibitem{selvaraju2017grad}
Selvaraju, R.R., Cogswell, M., Das, A., Vedantam, R., Parikh, D., Batra, D.: Grad-cam: Visual explanations from deep networks via gradient-based localization. In: Proceedings of the IEEE international conference on computer vision. pp. 618--626 (2017)

\bibitem{sohoni2020no}
Sohoni, N., Dunnmon, J., Angus, G., Gu, A., R{\'e}, C.: No subclass left behind: Fine-grained robustness in coarse-grained classification problems. Advances in Neural Information Processing Systems  \textbf{33},  19339--19352 (2020)

\bibitem{srivastava2020robustness}
Srivastava, M., Hashimoto, T., Liang, P.: Robustness to spurious correlations via human annotations. In: International Conference on Machine Learning. pp. 9109--9119. PMLR (2020)

\bibitem{taghanaki2021robust}
Taghanaki, S.A., Choi, K., Khasahmadi, A.H., Goyal, A.: Robust representation learning via perceptual similarity metrics. In: International Conference on Machine Learning. pp. 10043--10053. PMLR (2021)

\bibitem{wah2011caltech}
Wah, C., Branson, S., Welinder, P., Perona, P., Belongie, S.: The caltech-ucsd birds-200-2011 dataset  (2011)

\bibitem{wang2020score}
Wang, H., Wang, Z., Du, M., Yang, F., Zhang, Z., Ding, S., Mardziel, P., Hu, X.: Score-cam: Score-weighted visual explanations for convolutional neural networks. In: Proceedings of the IEEE/CVF conference on computer vision and pattern recognition workshops. pp. 24--25 (2020)

\bibitem{10.5555/3618408.3619982}
Wu, S., Yuksekgonul, M., Zhang, L., Zou, J.: Discover and cure: Concept-aware mitigation of spurious correlation. In: Proceedings of the 40th International Conference on Machine Learning. ICML'23, JMLR.org (2023)

\bibitem{xiao2020noise}
Xiao, K., Engstrom, L., Ilyas, A., Madry, A.: Noise or signal: The role of image backgrounds in object recognition. ArXiv preprint arXiv:2006.09994  (2020)

\bibitem{xuan2024attributionscanner}
Xuan, X., Ono, J.P., Gou, L., Ma, K.L., Ren, L.: Attributionscanner: A visual analytics system for metadata-free data-slicing based model validation. arXiv preprint arXiv:2401.06462  (2024)

\bibitem{xue2023eliminating}
Xue, Y., Payani, A., Yang, Y., Mirzasoleiman, B.: Eliminating spurious correlations from pre-trained models via data mixing. arXiv preprint arXiv:2305.14521  (2023)

\bibitem{yan2023towards}
Yan, S., et~al.: Towards trustable skin cancer diagnosis via rewriting model's decision. In: IEEE CVPR (2023)

\bibitem{yang2023identifying}
Yang, Y., Gan, E., Karolina~Dziugaite, G., Mirzasoleiman, B.: Identifying spurious biases early in training through the lens of simplicity bias. In: Proceedings of The 27th International Conference on Artificial Intelligence and Statistics. Proceedings of Machine Learning Research, vol.~238, pp. 2953--2961. PMLR (02--04 May 2024)

\bibitem{10.5555/3618408.3620051}
Yang, Y., Nushi, B., Palangi, H., Mirzasoleiman, B.: Mitigating spurious correlations in multi-modal models during fine-tuning. In: Proceedings of the 40th International Conference on Machine Learning. ICML'23, JMLR.org (2023)

\bibitem{zhang2022correct}
Zhang, M., Sohoni, N.S., Zhang, H.R., Finn, C., R{\'e}, C.: Correct-n-contrast: A contrastive approach for improving robustness to spurious correlations. arXiv preprint arXiv:2203.01517  (2022)

\bibitem{zhang2022nico}
Zhang, X., He, Y., Xu, R., Yu, H., Shen, Z., Cui, P.: Nico++: Towards better benchmarking for domain generalization. In: Proceedings of the IEEE/CVF Conference on Computer Vision and Pattern Recognition. pp. 16036--16047 (2023)

\bibitem{zhou2017places}
Zhou, B., Lapedriza, A., Khosla, A., Oliva, A., Torralba, A.: Places: A 10 million image database for scene recognition. IEEE transactions on pattern analysis and machine intelligence  \textbf{40}(6),  1452--1464 (2017)

\bibitem{zhou2003learning}
Zhou, D., Bousquet, O., Lal, T., Weston, J., Sch{\"o}lkopf, B.: Learning with local and global consistency. Advances in neural information processing systems  \textbf{16} (2003)

\end{thebibliography}

\newpage
\appendix
\setcounter{equation}{0}
\renewcommand\theequation{A\arabic{equation}} 
\setcounter{table}{0}  
\setcounter{figure}{0}
\renewcommand{\thetable}{A\arabic{table}}
\renewcommand{\thefigure}{A\arabic{figure}}

\section{Appendix}\label{sec:appx}

\noindent This appendix section is organized as follows:
\begin{itemize}
    \item Sec.~\ref{sec:Proof Prelim} presents the preliminary knowledge underpinning our interpretation in Section~\ref{appendix:proof_lemma}.
    \item Sec.~\ref{appendix:proof_lemma} presents a theoretical interpretation to further support our choice of preserving data that receives correct attention to enhance core feature learning.
    \item Sec.~\ref{sec:proof_aux} presents the theorem that grounds our interpretation in Sec.~\ref{appendix:proof_lemma}.
    \item Sec.~\ref{appendix:datasets} provides a more detailed description of the datasets we used for the experiments.
    \item Sec.~\ref{appendix:Training} presents the training setting for our experiments.
    \item Sec.~\ref{appendix:Crowd} provides the interface and instructions for our crowdsourcing tasks.
    \item Sec.~\ref{appendix:AIOU} provides the definition of AIoU score.
    \item Sec.~\ref{appendix:More Results} offers more examples to evaluate attention consistency in the original feature space, our constructed attention space, and the environmental feature spaces. Additionally, we provide further examples to validate \slim's enhanced attention accuracy.
\end{itemize}

\subsection{Proof Preliminaries}\label{sec:Proof Prelim}
In this appendix, we use lowercase letters, lowercase boldface letters, and uppercase boldface letters to respectively denote scalars (a), vectors ($\textbf{v}$), and matrices ($\textbf{W}$).

To simplify the complex real-world issue of spurious correlations into a formal framework, in alignment with previous works~\cite{chen2022towards,jelassi2022towards,deng2023robust}, we adopt a two-layer nonlinear convolutional neural network (CNN) based on a data model that captures the influence of spurious correlations. 
The two-layer nonlinear CNN is defined as follows:
\begin{equation}\label{eqn:cnn}
    f(\textbf{x};\textbf{W}) = \sum_{j\in [J]} \sum_{p=1}^P \sigma(\langle \text{w}_j,\text{x}^{(p)} \rangle),
\end{equation}
where $\textbf{w}_j\in \mathbb{R}^d$ is the weight vector of the $j$-th filter, $J$ is the number of filters (neurons) of the network, and $\sigma(z)=z^3$ is the activation function. 
$\textbf{W}=[\textbf{w}_1,\dots, \textbf{w}_J]\in \mathbb{R}^{d\times J}$ denotes the weight matrix of the CNN.
In\cite{deng2023robust,jelassi2022towards,cao2022benign}, they assume a mild overparameterization of the CNN with $J = polylog(d)$ and initialize $\textbf{W}^{(0)} \sim N (0,\sigma^2_0)$, where $=polylog(d)/d$. 

To understand the underlying dynamics in feature learning, we introduce the following data model where the input consists of a core feature, a spurious feature, and noise patches. 
\begin{definition}[Data model.\cite{deng2023robust}]\label{def:data model}
A data point $(\textbf{x}, y, s)\in (\mathbb{R}^d)^P\times \{\pm 1\}\times \{\pm 1\}$ is generated from the distribution $\mathcal D$ as follows.
\begin{itemize}
    \item Randomly generate the true label $y\in \{\pm 1\}$.
    \item Generate spuriousness label $s\in \{\pm y\}$, where $s=y$ with probability $\alpha > 0.5$.
    \item Generate $\textbf{x}$ as a collection of $P$ patches: $\textbf{x}=(\textbf{x}^{(1)},\textbf{x}^{(2)},\dots,\textbf{x}^{(P)})\in (\mathbb{R}^d)^P$, where
   
    \begin{itemize}
    \item Core Feature. One and only one patch is given by $\beta_c \cdot y \cdot \textbf{v}_c$ with $||\textbf{v}_c||_2=1$. $\beta_c$ is the core feature strength.
    \item Spurious Feature. One and only one patch is given by $\beta_s \cdot s \cdot \textbf{v}_s$ with $||\textbf{v}_s||_2=1$ and $ \langle \textbf{v}_c,\textbf{v}_c \rangle = 0$. $\beta_s$ is the spurious feature strength.
    \item Random noise. The rest of the $P-2$ patches are Gaussian noises $\xi$ that are independently drawn from $\textit{N}(0,(\sigma^2_p/d) \cdot \textbf{I}_d)$ with $\sigma_p$ as an absolute constant.
    \end{itemize}
\end{itemize}
\end{definition}

\noindent With the given data model, considering the training dataset $S = \{(\textbf{x}_i, y_i, a_i)\}^N_{i=1}$ and let $S$ be partitioned into large group $S_1$ and small group $S_2$ such that $S_1$ contains all the data that can be correctly classified by the spurious feature, i.e., $s_i = y_i$, and $S_2$ contains all the data that can only be correctly classified by the core feature, i.e., $s_i = -y_i$. Denote $\hat{\alpha} = \frac{|S_1|}{N}$ and therefore $1-\hat{\alpha} = \frac{|S_2|}{N}$.

\noindent\textbf{Remark.} Different from the original definition in\cite{deng2023robust}, we do not make assumptions about the relative strengths of $\beta_c$ and $\beta_s$. 
Rather, our approach estimates the relative strengths of $\beta_c$ and $\beta_s$ through attention correctness annotations, as the saliency map can reflect the features learned by the model.

\subsection{Theoretical Inspiration}\label{appendix:proof_lemma}

In Sec.~\textcolor{red}{5}, we have demonstrated the robustness of annotating attention correctness and corroborated that decoupling core and environment features is crucial for learning core features.
In building feature-balanced datasets, our approach primarily focuses on leveraging data that receives correct attention from the reference model.
One reason for this is that environment features can be more accurately and efficiently isolated based on the identified core features.
Adopting the \textbf{Theorem}~\ref{thm:bound} proposed in \cite{deng2023robust}, we interpret it from a different perspective to further support and justify that preserving data with high attention scores guarantees the effective learning of core features in a more balanced dataset.

\begin{lemma}\label{lma:diff}

Under training dataset $S$, which follows the distribution described in \textbf{Definition}~\ref{def:data model}, when the data is trained using gradient descent for $T_0 = \tilde{\mathrm{\Theta}}(\eta)(1/\eta\beta_s^3\sigma_0)$ iterations on the model as introduced in Eqn.~\ref{eqn:cnn}, instances receiving higher attention scores are more likely to have their core features learned in a new training scenario with a more balanced data distribution (i.e., $\hat{\alpha}\rightarrow 1/2$).
\end{lemma}

\begin{proof}
Based on \textbf{Theorem}~\ref{thm:bound}, we are implied that in the early $T_0$ iterations
\begin{equation}\label{eqn:learn_spur}
    \beta_c \ll \beta_s \sqrt[3]{2\hat{\alpha}-1} \Rightarrow P_{lrc}\rightarrow 0,
\end{equation}
where $P_{lrc}$ is the probability of the model learned the core feature. Thus, the converse-negative proposition of proposition~(\ref{eqn:learn_spur}) is 
\begin{equation}\label{eqn:learn_core}
     P_{lrc} > 0 \Rightarrow \beta_c \not\ll \beta_s \sqrt[3]{2\hat{\alpha}-1},
\end{equation}
in the early $T_0$ iterations. 
For instance $\textbf{x}_i$, whose core feature has been effectively learned, there should exist a constant threshold, denoted as $Tr(\textbf{x}_i)$. 
This threshold ensures that the core feature has the chance to be learned once $\beta_c - \beta_s \sqrt[3]{2\hat{\alpha}-1} > Tr(\textbf{x}_i)$. 
Intuitively, $P_{lrc}(\textbf{x}_i) \propto (\beta_c(\textbf{x}_i) - \beta_s(\textbf{x}_i) \sqrt[3]{2\hat{\alpha}-1})$.
Since the strengths of the core feature $\beta_c(\textbf{x}_i)$ and spurious feature $\beta_s(\textbf{x}_i)$ are natures of the data itself and do not change, in a more balanced data distribution, as $\hat{\alpha} \rightarrow 1/2$, $(\beta_c(\textbf{x}_i) - \beta_s(\textbf{x}_i) \sqrt[3]{2\hat{\alpha}-1})$ is increasing, consequently $P_{lrc}(\textbf{x}_i)$ is increasing.
Since the learned feature of an instance can be interpreted via saliency maps, a higher attention score means that its core feature has been learned more accurately. 
Therefore, such instances will have a higher probability of the core feature being continuously learned as the data distribution becomes more balanced.
\qed
\end{proof}
Although this theoretical insight is built on a simplified binary classification model, it provides an inspirational hint towards understanding the benefit of utilizing data with high attention scores in more complex scenarios.

\subsection{Auxiliary Theorem}\label{sec:proof_aux}

\begin{theorem}\label{thm:bound} (Theorem 2.2 in\cite{deng2023robust}.) Consider the training dataset $S$ that follows the distribution in Definition~\ref{def:data model}. 
Consider the two-layer nonlinear CNN model as in Eqn.~(\ref{eqn:cnn}) initialized with $\textbf{W}^{(0)} \sim N (0,\sigma^2_0)$. 
After training with gradient decent for $T_0 = \tilde{\mathrm{\Theta}}(1/\eta\beta_s^3\sigma_0)$ iterations, for all $j \in [J]$ and $t \in [0, T_0)$, we have
\begin{align}\label{eqn:bound}
    \tilde{\mathrm{\Theta}}(\eta)\beta_s^3(2\hat{\alpha}-1) \langle \textbf{w}_j^{(t)},\textbf{v}_s \rangle^2 \le \langle \textbf{w}_j^{(t+1)},\textbf{v}_s \rangle - \langle \textbf{w}_j^{(t)},\textbf{v}_s \rangle  \le \tilde{\mathrm{\Theta}}(\eta)\beta_s^3 \hat{\alpha} \langle \textbf{w}_j^{(t)},\textbf{v}_s \rangle^2,\\
     \tilde{\mathrm{\Theta}}(\eta)\beta_c^3\hat{\alpha} \langle \textbf{w}_j^{(t)},\textbf{v}_c \rangle^2 \le \langle \textbf{w}_j^{(t+1)},\textbf{v}_c \rangle - \langle \textbf{w}_j^{(t)},\textbf{v}_c \rangle  \le \tilde{\mathrm{\Theta}}(\eta)\beta_c^3  \langle \textbf{w}_j^{(t)},\textbf{v}_c \rangle^2.
\end{align}
\end{theorem}
\noindent With the updates of the spurious and core feature in the early iterations, \textbf{Theorem}~\ref{thm:bound} gives the condition-if $\beta_c^3<\beta_s^3(2\hat{\alpha}-1)$-that GD will learn the spurious feature very quickly while hardly learning the core feature.

\subsection{Datasets}\label{appendix:datasets}
\noindent\textbf{Waterbirds~\cite{Sagawa2020Distributionally}.} It is constructed to study the spurious correlation between the image background and the object. 
To this end, bird images in Caltech-UCSD Birds-200-2011 (CUB-200-2011) dataset~\cite{wah2011caltech} are grouped into waterbirds and landbirds. All birds are then cut and pasted onto new background images from the Places dataset~\cite{zhou2017places}, with waterbirds having a higher probability on water and landbirds having a higher probability on land.
The training set contains 4,795 images in total, 3,498 for landbirds with land background, 184 for landbirds with water background, 56 for waterbirds with land background, and 1,057 for waterbirds with water background.
The validation set contains 1,199 images in total, 467 for landbirds with land background, 466 for landbirds with water background, 133 for waterbirds with land background, and 133 for waterbirds with water background.

\noindent\textbf{CelebA~\cite{liu2015deep}.} It is a large-scale face attribute dataset comprised of photos of celebrities. Each image is annotated with 40 binary attributes. Aligned with other works focusing on spuriousness mitigation, we chose ``blond hair'' or ``non-blond hair'' as the target attributes, and gender as the spurious feature for hair color classification.
The training set contains 162,770 images in total, 71,629 for non-blond haired female, 66,874 for non-blond haired male, 22,880 for blond haired female, and 1,387 for blond haired male.
The validation set contains 19,867 images in total, 8,535 for non-blond haired female, 8,276 for non-blond haired male, 2,874 for blond haired female, and 182 for blond haired male.

\noindent\textbf{ISIC~\cite{8363547}.} It contains images of a skin lesion, categorized into (1) benign lesions or (2) malignant lesions. In a real-life task, this would be done to determine whether a biopsy should be taken.
Aligning with previous studies\cite{10.5555/3524938.3525689}, we target colorful patches as spurious features, and also follow the same strategy to obtain data from its official platform.

\begin{figure*}[thbp]
  \centering
   \includegraphics[width=0.6\linewidth]{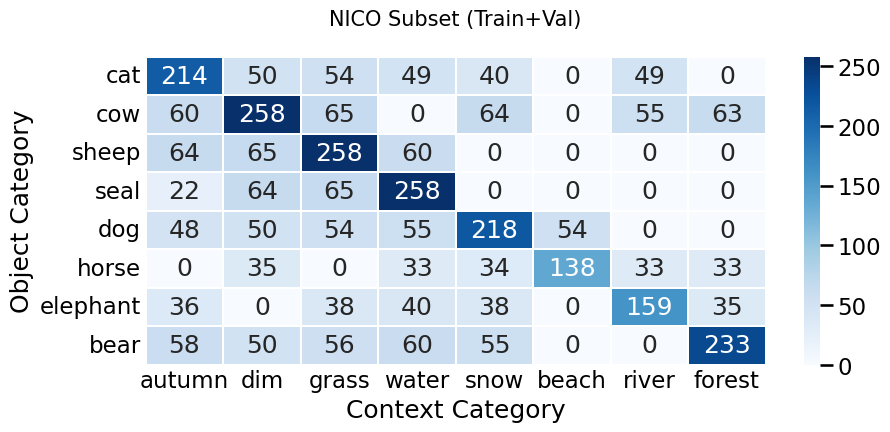}
\caption{Data distribution in the combination of training and validation set of NICO, with respect to object and context categories.}
   \label{fig:nico_dis}
\end{figure*}

\noindent\textbf{NICO.} 
Derived from NICO++\cite{zhang2022nico}, this dataset features various object categories in shifted contexts to probe spurious correlations.
It is a multi-class image dataset presenting a diverse set of objects in varied contextual scenarios, allowing convenient adjustment of the distributions of object and context labels. We randomly sample eight animal categories with eight different contextual labels. To challenge the model with spurious correlations, the training set distribution follows three rules: (a) each object class is distributed across various contexts; (b) there is one dominant context for each single class; (c) the dominant context for each class is unique. For instance, most ``sheep'' images have context ``grass'' and ``grass'' context is only dominant in ``sheep''. The detailed distribution is shown in Fig.~\ref{fig:nico_dis}.

\noindent\textbf{ImagetNet9 (IN9)~\cite{xiao2020noise}.}
This dataset is a curated subset extracted from the larger ImageNet collection, specifically designed to scrutinize and address the model bias towards object backgrounds. To evaluate our framework for spurious correlation mitigation, we adopt the ``Mixed-Rand'' setup, which is a particular data arrangement where images are organized to have a randomized correlation between the object and its background. This setup aims to challenge models to focus on the object itself rather than the background, helping to test and improve the robustness of models against spurious correlations.
We utilize the training and validation splits provided by ImageNet9, ensuring that our experiments are aligned with established benchmarks for consistency and comparability. Moreover, their provided model trained on IN-9L is used as our reference model. The outcomes of our experiments are then evaluated on the Mixed-Rand.

\subsection{Training Setting}\label{appendix:Training}
In Table \textcolor{red}{4}, we present the amount of data required for attention annotation and the size of the constructed data used for model training. In this section, we provide additional details on the training settings.
To maintain consistency with existing state-of-the-art methods, we use SGD as the optimization algorithm. The ranges of hyperparameters, batch sizes, and total training epochs employed in our experiments are all tuned based on these methods~\cite{Sagawa2020Distributionally,zhang2022correct,kirichenko2022last,deng2023robust,yang2023identifying}, as listed in Table~\ref{tab:hyper}.
The detailed training setups corresponding \slimVal are following DFR~\cite{kirichenko2022last}.
All experiments are conducted on two NVIDIA RTX 4090 GPUs with 24GB memory.

\begin{table}[htbp]
\caption{Hyperparameters used for the \slimTr's results in Sec.~\textcolor{red}{5.2} on different datasets.}
\label{tab:hyper}
\resizebox{\columnwidth}{!}{%
\begin{tabular}{p{2.5cm}<{\centering}p{2.2cm}<{\centering}p{2.2cm}<{\centering}p{2.2cm}<{\centering}p{2.2cm}<{\centering}p{2.2cm}<{\centering}}
\toprule[0.8pt]
Dataset & Waterbirds & CelebA & ISIC & NICO & ImageNet 9 \\ \hline
Initial lr            & 1E-3                              & 1E-4             &  0.002           & 1E-6            &  1E-6                         \\
Weight Decay             & 0.1               & 0.1              &  0.1            & 0.5            &  0.1                        \\
Batch Size               & 128               &128               & 128             & 128            &128                       \\
Training Epochs         & 50               & 30              & 30             &  50           &  10                        \\
Core Cluster             & 2               & 2              & 2             &  8           &  9                       \\
Env Cluster              & 3               & 3             & 2            & 10            & 9                         \\ \bottomrule[0.8pt]
\end{tabular}%
}
\end{table}

\subsection{Crowdsourcing Instruction}\label{appendix:Crowd}
In Sec.~\textcolor{red}{5.3}, we employed crowdsourcing tasks with the Waterbirds and NICO datasets to compare the consistency of annotating spuriousness versus attention correctness. For this study, we selected a random set of 120 images from each dataset and established two separate tasks: one for spuriousness labeling and another for attention correctness labeling. For each task, we recruited 60 participants who are native English speakers, independently from the Prolific platform, to prevent learning biases from cross-task participation. Participants received an hourly fee for their participation. In ensuring ethical research standards, our study refrained from collecting personally identifiable information and excluded any potentially offensive content.

We provided the following instruction to the participants for the spuriousness labeling task: ``This study focuses on the evaluation of the image annotation tasks.
Participants will be presented with a series of $***$ images featuring different animals. For each image, the task involves selecting one most accurate description of the primary background from provided options.
There are no specific prerequisites for participation. Simply make selections based on your observation. 
Notice: This study ensures the confidentiality of your participation, as it neither collects personally identifiable information nor contains any offensive content. Your feedback will exclusively be utilized for academic research purposes.''

We provided the following instruction to the participants for the attention correctness labeling task: ``This study aims to evaluate a Machine Learning Model's attention correctness.
Participants will review $***$ image pairs. Each pair includes an original bird image and a version with a highlighted overlay indicating the model's focus area.
The task is to choose the more accurate description of the highlighted region from two options.
No special skills are required for participation. Simply make selections based on your observation. 
Notice: This study ensures the confidentiality of your participation, as it neither collects personally identifiable information nor contains any offensive content. Your feedback will exclusively be utilized for academic research purposes.''

\begin{figure*}[htbp]
  \centering
   \includegraphics[width=1\linewidth]{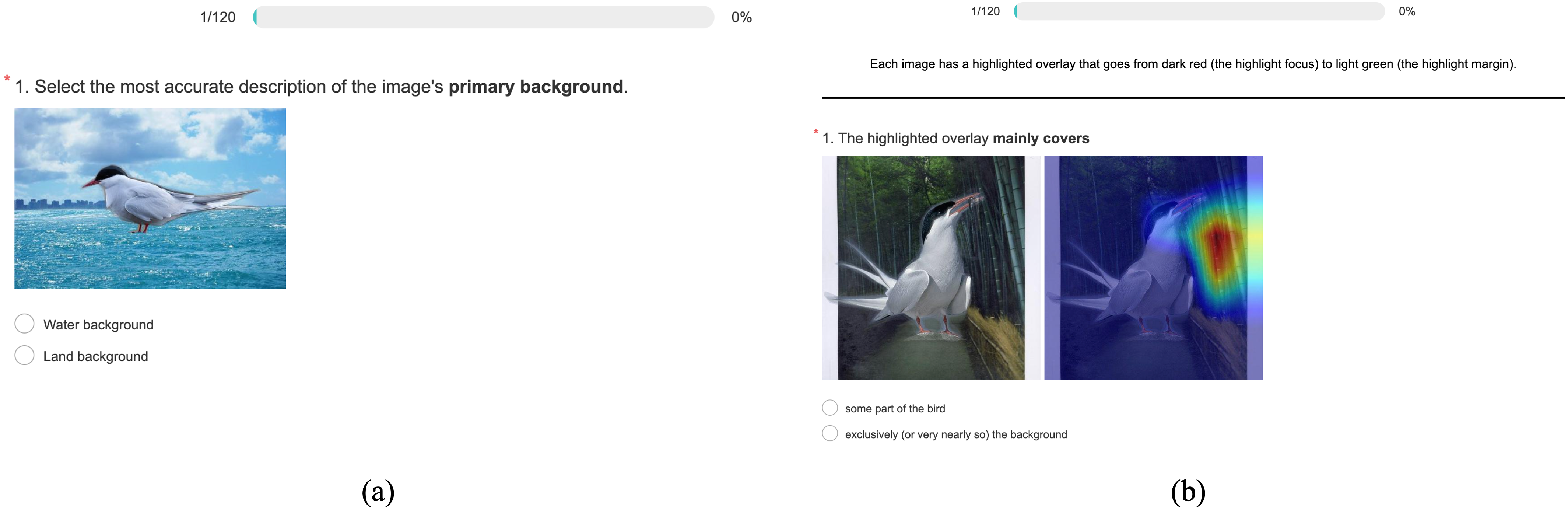}
\caption{Crowdsourcing interface for (a) spuriousness labeling and (b) attention correctness annotation on Waterbirds dataset.
}
   \label{fig:crowdW}
\end{figure*}

\begin{figure*}[htbp]
  \centering
   \includegraphics[width=1\linewidth]{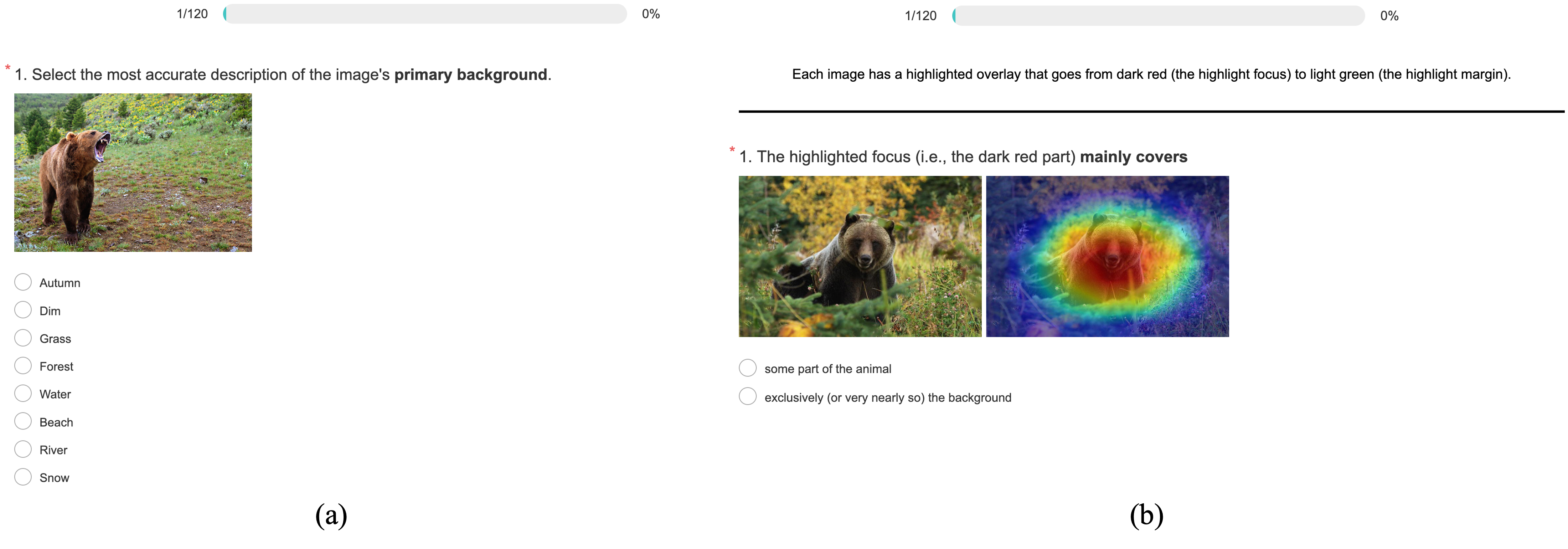}
\caption{Crowdsourcing interface for (a) spuriousness labeling and (b) attention correctness annotation on NICO dataset.
}
   \label{fig:crowdN}
\end{figure*}

The interfaces for spuriousness labeling task are listed in Fig.~\ref{fig:crowdW}.(a) and Fig.~\ref{fig:crowdN}.(a).
And the interfaces for attention correctness labeling task are listed in Fig.~\ref{fig:crowdW}.(b) and Fig.~\ref{fig:crowdN}.(b).

For the results provided in Secs.~\textcolor{red}{5.2} and~\textcolor{red}{5.4}, we utilized a similar attention correctness labeling instruction and interface.
The differences are as follows: (1) the instances selected for annotation are based on our proposed sampling strategy, as introduced in Sec.~\textcolor{red}{4.2}; (2) in the attention correctness labeling interface, option (a) is ``some part of the $\{*\}$,'' where $\{*\}$ represents the specific prediction corresponding to the image. 
In the case of ISIC dataset, we collaborated with domain experts to obtain annotation. For the other datasets, our participants were sourced from the Prolific platform.

\subsection{AIoU}\label{appendix:AIOU}

Previous research has often employed binary attribute maps to calculate the Intersection-over-Union (IoU) score against the ground-truth bounding box\cite{nguyen2021the}. 
\begin{equation}\label{eqn:IOU}
IoU(M,B) = \frac{\sum_{j,k}\min(M_{jk},B_{jk})}{\sum_{j,k}\max(M_{jk},B_{jk})},
\end{equation}
However, the conventional IoU's reliability for assessing attribute map quality is compromised by its sensitivity to the chosen binarization threshold. To overcome this limitation, the revised approach\cite{10.5555/3618408.3620051} replaces the binary intersection with a minimum operator between a bounding box $B_y$ and an explanation map $M_y$ of ground truth calss $y$. AIoU employs a maximum operator in place of the binary union, facilitating a more consistent evaluation that is less susceptible to thresholding variations.
Eqn.(\ref{eqn:IOU}) assesses the alignment between an explanation map and the ground-truth bounding box; however, it overlooks the possibility that, despite precise alignment for the correct class, explanation maps for alternative classes might overlap with the bounding box of the true class. 
\begin{equation}\label{eqn:AIOU}
AIoU = \frac{IoU(M_y,B_y)}{IoU(M_y,B_y)+\max_{y\prime \in [C/ y]}IoU(M_{y\prime},B_y)},
\end{equation}
Consequently, AIoU is a modified IoU metric that refines its denominator to account for the class with the explanation map exhibiting the maximum intersection with the ground-truth bounding box. In our evaluation, we use {\small GradCAM} as the explanation map.

\subsection{Additional Experiment Results}\label{appendix:More Results}
\begin{table}[t]
\caption{Ablation study on the dimensionality of attention space. } 
\centering
\resizebox{0.95\columnwidth}{!}{%
\begin{tabular}{p{3cm}<{\centering}p{1.8cm}<{\centering}p{1.8cm}<{\centering}p{1.8cm}<{\centering}p{1.8cm}<{\centering}}
\toprule[0.8pt]
Annotation Amounts & \texttt{Dim} $=2$ & \texttt{Dim} $=3$ &\texttt{Dim} $=5$ & \texttt{Dim} $=10$ \\ \hline
60 (1.3\%)   & $78.21_{\pm 0.52}$      & $78.43_{\pm 0.44}$     &  $78.91_{\pm 0.41}$    &   $79.73_{\pm 0.43}$    \\
90 (1.9\%)   & $85.93_{\pm 0.92}$      & $86.24_{\pm 0.96}$     &  $86.18_{\pm 0.91}$    &   $86.20_{\pm 0.94}$    \\
120 (2.5\%)  & $89.12_{\pm 0.64}$      & $89.12_{\pm 0.72}$     &  $89.13_{\pm 0.72}$    &   $89.12_{\pm 0.74}$   \\ \bottomrule[0.8pt] 
\end{tabular}%
} 
\label{tab:ablation_2d}
\end{table}

\begin{table}[ht]
\caption{Results with additional model architectures and datasets. } 
\centering
\resizebox{0.95\columnwidth}{!}{%
\begin{tabular}{p{1.3cm}<{\centering}p{1.5cm}<{\centering}p{1.5cm}<{\centering}p{1.5cm}<{\centering}p{1.5cm}<{\centering}p{1.5cm}<{\centering}p{1.5cm}<{\centering}}
\toprule[0.5pt]
    & \multicolumn{2}{c}{Waterbirds (ViT)}  & \multicolumn{2}{c}{MetaShift (ResNet50)}  & \multicolumn{2}{c}{FMoW (DenseNet121)} 
    \\ 
    \cmidrule(r){2-3} \cmidrule(r){4-5}  \cmidrule(r){6-7}
\multirow{-2}{*}{Method} & Worst           & Avg          & Worst         & Avg          & Worst         & Avg  \\ \hline
ERM                                &  $85.5_{\pm 1.2}$               & $96.3_{\pm 0.5}$              & $62.1_{\pm 4.8}$              &  $72.9_{\pm 1.4}$    & $32.3_{\pm 1.3}$              &  $53.0_{\pm 0.6}$                 \\


JTT                           & $86.7_{\pm 1.5}$                & $95.3_{\pm 0.7}$              & $64.6_{\pm 2.3}$              &  $74.4_{\pm 0.6}$   &$33.4_{\pm 0.9}$              &  $52.5_{\pm 0.3}$               \\
\hline
DISC                     & \underline{$91.5_{\pm 1.3}$}                   &  $95.3_{\pm 1.1}$            &  \underline{$73.5_{\pm 1.4}$}             & $75.5_{\pm 1.1}$          &\underline{$36.1_{\pm 1.8}$}             & $53.9_{\pm 0.4}$       \\

\rowcolor[HTML]{EFEFEF} 
\slimTr         &  \bm{$92.1_{\pm 0.6}$}              & $96.4_{\pm 0.3}$   &  \bm{$75.7_{\pm 1.0}$}             & $76.4_{\pm 0.8}$             & \bm{$37.4_{\pm 1.1}$}             & $54.1_{\pm 0.4}$\\ 

\hline
GDRO             &  $91.3_{\pm 0.8}$               &  $94.9_{\pm 0.3}$      &    $66.0_{\pm 3.8}$             & $73.6_{\pm 2.1}$   &    $30.8_{\pm 0.8}$             & $52.1_{\pm 0.5}$     \\ 

\bottomrule[0.5pt]
\end{tabular}%
}
\label{tab:reb_performance}
\end{table}

\noindent\textbf{Influence of the attention space’s
dimension.} Table~\ref{tab:ablation_2d} shows \slimTr's ablation study results on Waterbirds, where for each annotation amount ($N$), we only vary the attention space's \texttt{dim} and measure \texttt{worst-group acc}. 
Results reveal a slight performance improvement with higher \texttt{dims} when $N$=$60$, but it is much less than the performance boosting caused by increasing $N$. 
When $N$=$120$ (same as Table~\textcolor{red}{2} setting), we observe stable performance when varying \texttt{dims}. 
As $120$ is a modest annotation amount, 2 \texttt{dim} is preferred.

\noindent \textbf{Results with additional model architectures and datasets.}
Table~\ref{tab:reb_performance} includes results: (1) on Waterbirds using a reference model matching ViT-S/16 in \cite{ghosal2024vision}; and (2) on MetaShift and FMoW, using reference models matching the corresponding ones in DISC~\cite{10.5555/3618408.3619982}.
Table~\ref{tab:reb_performance} again confirms \slim{}'s outstanding performance over the baselines. 

\begin{figure*}[htbp]
  \centering
   \includegraphics[width=1\columnwidth]{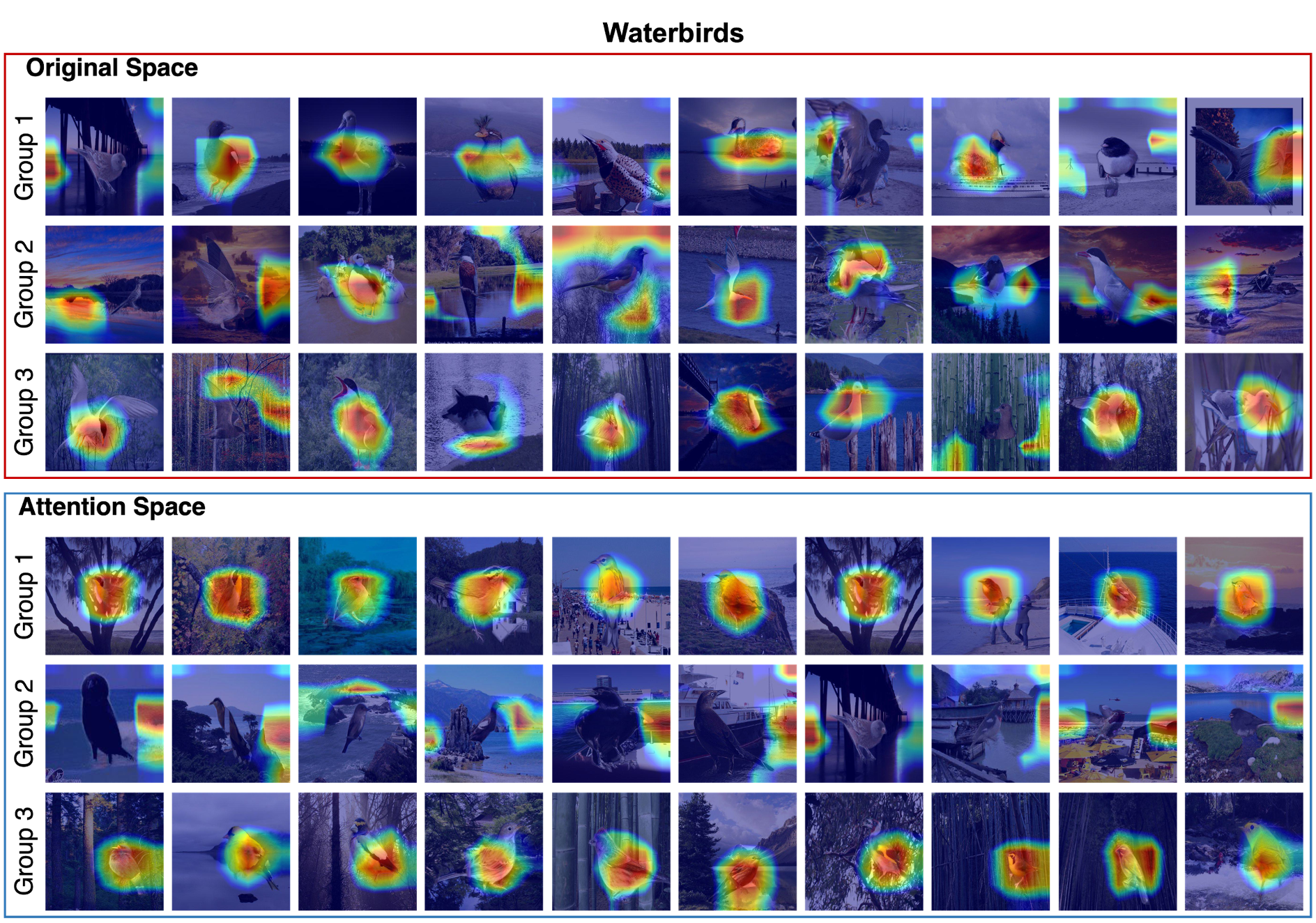}
\caption{Comparison of attention consistency between the original representation and attention spaces on the Waterbirds dataset. Examples in each group represent nearest neighbors within the corresponding space.}
\label{fig:AttConsistencyW}
\end{figure*}

\begin{figure*}[htbp]
  \centering
   \includegraphics[width=1\columnwidth]{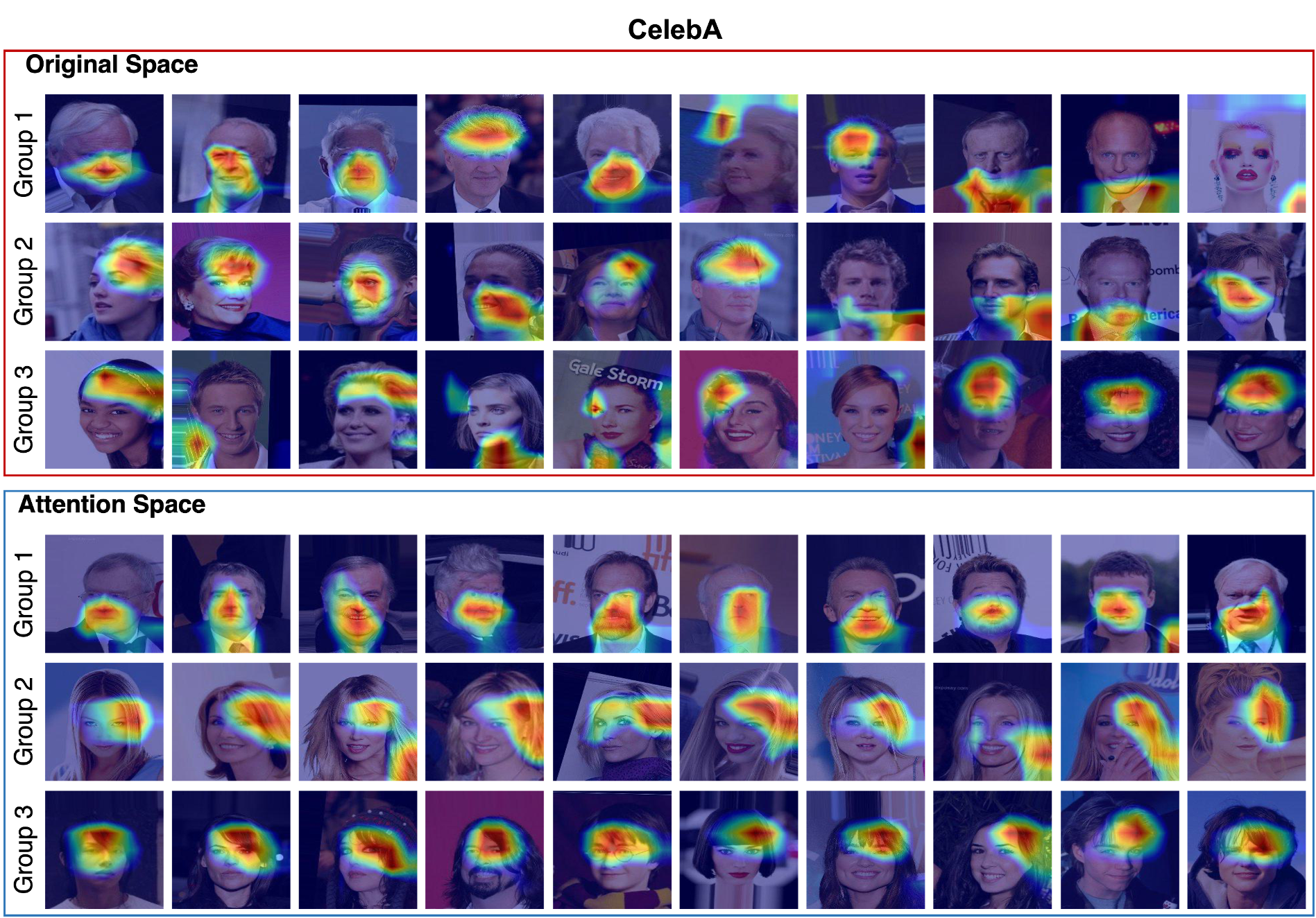}
\caption{Comparison of attention consistency between the original representation and attention spaces on the CelebA dataset. Examples in each group represent nearest neighbors within the corresponding space.}
\label{fig:AttConsistencyC}
\end{figure*}

\noindent\textbf{Attention Consistency.}
In Sec.~\textcolor{red}{5.3}, we have quantitatively compared how similar neighbors are in the original feature space versus the attention space created by \slim. 
In this section, we provide qualitative comparison by randomly selecting three points and examining the GradCAMs of their 10 nearest neighbors in the original representation space and our proposed attention space as showcased in Figs.~\ref{fig:AttConsistencyW} and \ref{fig:AttConsistencyC}.
We can observe that, unlike the original space, the attention space aptly groups instances with coherent attributions. This facilitates the attention annotation and expansion with consistent attribution patterns.

\begin{figure*}[htbp]
  \centering
   \includegraphics[width=1\columnwidth]{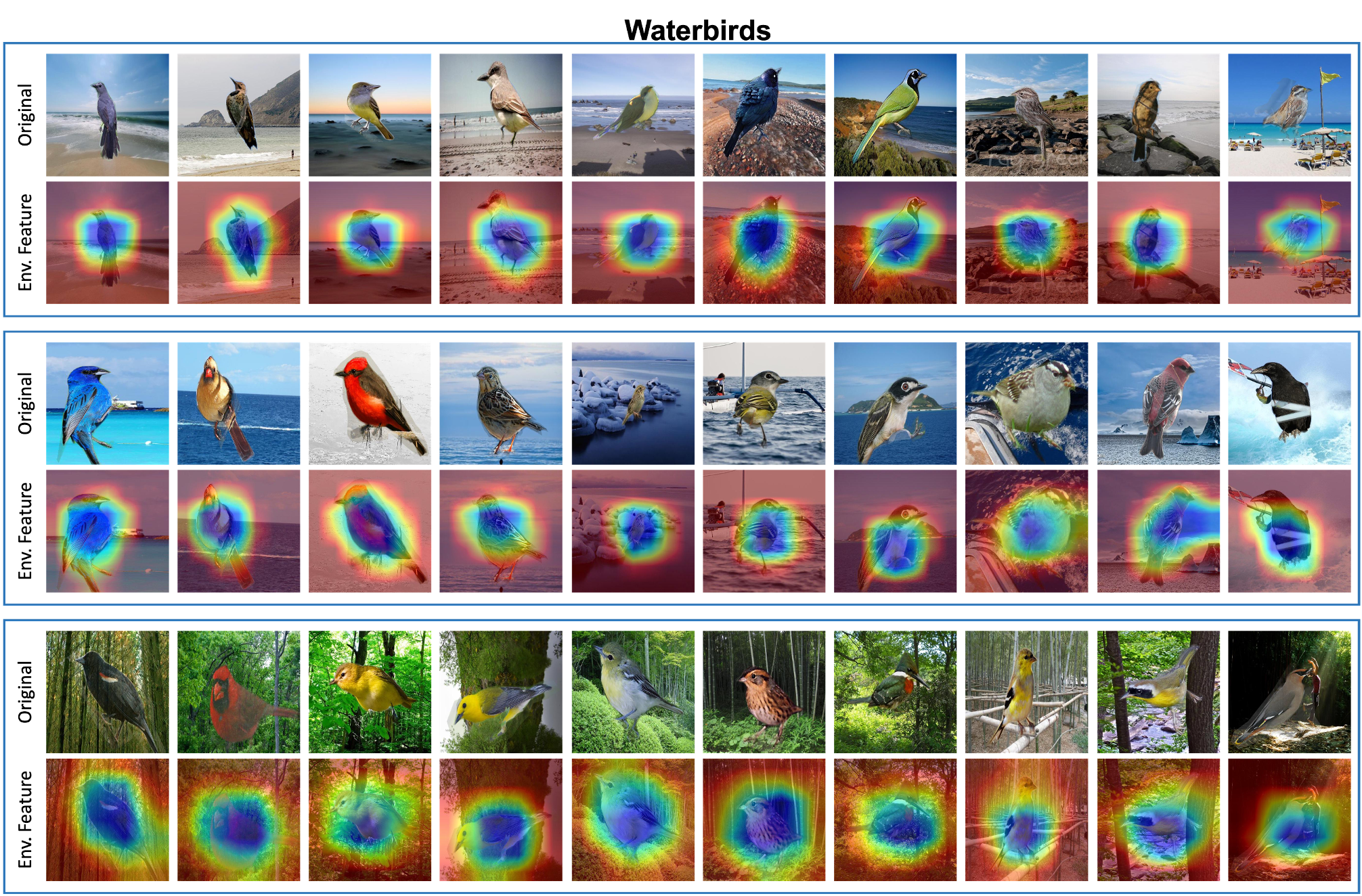}
\caption{Three groups of examples from the environment feature space on the Waterbird dataset.}
\label{fig:envconsisW}
\end{figure*}

\begin{figure*}[htbp]
  \centering
   \includegraphics[width=1\columnwidth]{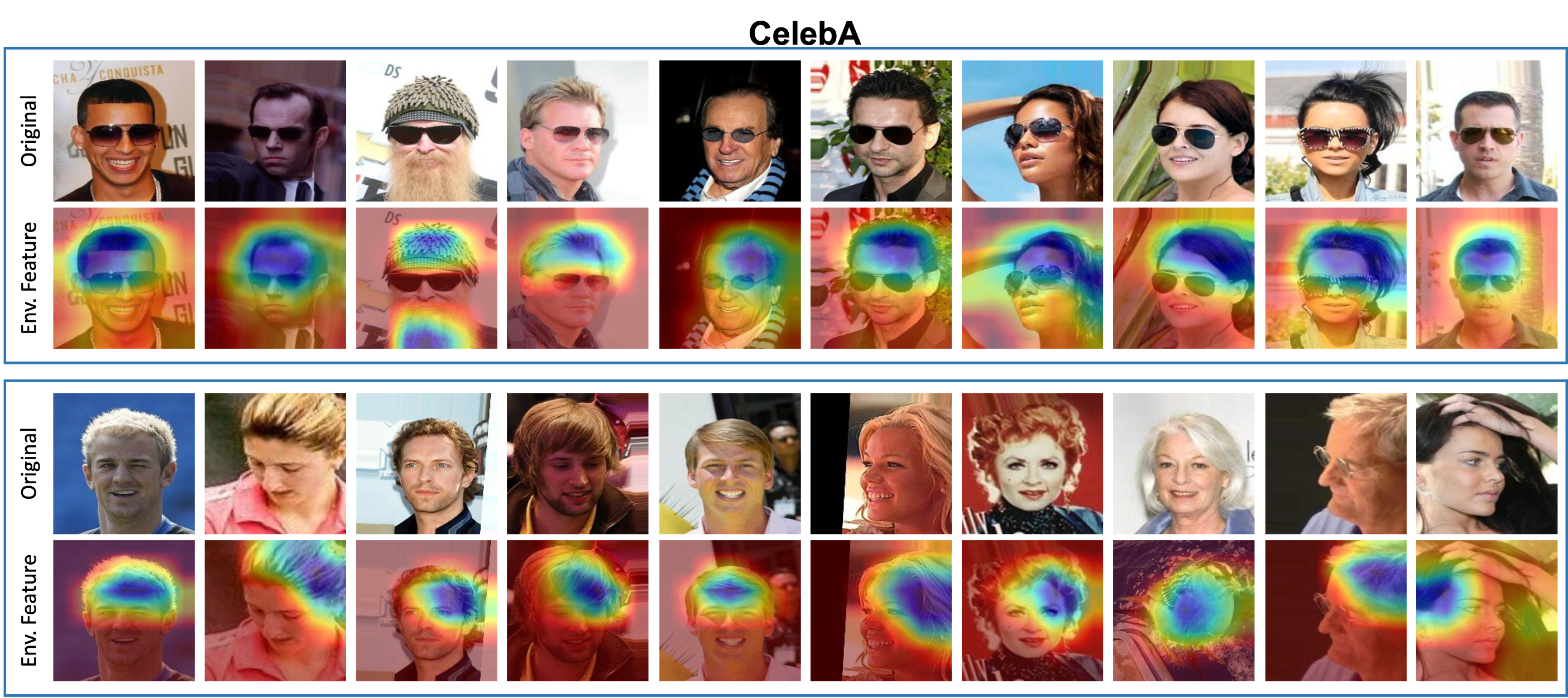}
\caption{Two groups of examples from the environment feature space on the CelebA dataset.}
\label{fig:envconsisC}
\end{figure*}

\noindent\textbf{Environment Feature Space.}
After disentangling core and environment features, we construct environment feature sets based on the inverse-attention-weighted features vectors $F_{\hat{A}}$. 
Here, we provide some intuitive examples to verify the consistency of the environment feature within clusters and the diversity of the environment feature between clusters in the environment feature space.
We randomly select points from different clusters in environment feature space and examine the GradCAMs of their 10 nearest neighbors, the results as showcased in Figs.~\ref{fig:envconsisW} and~\ref{fig:envconsisC}.
As illustrated in Fig.~\ref{fig:envconsisW}, each group of data has a high consistency in environment features, such as land and sea backgrounds, ocean backgrounds, and forest backgrounds. 
This example demonstrates that we can effectively estimate the environment features by weighting $F$ with inverse attention masks $\bar{A}$ (visualized as GradCAMs in Fig.~\ref{fig:envconsisW}) after identifying the core attention mask $A$.
Furthermore, we find that compared to manually labeling spurious features, this proposed method allows us to identify different types of environment features more accurately and in greater detail.
This paves the way for our ultimate goal: ensuring a balanced representation of core features across various environment features.
In Fig.~\ref{fig:envconsisC}, we observe a similar situation: after isolating the core feature, namely hair, the first group exhibits a consistent environment feature, such as wearing glasses, while the second group consistently appears as white individuals.

\begin{figure*}[htbp]
  \centering
   \includegraphics[width=1\columnwidth]{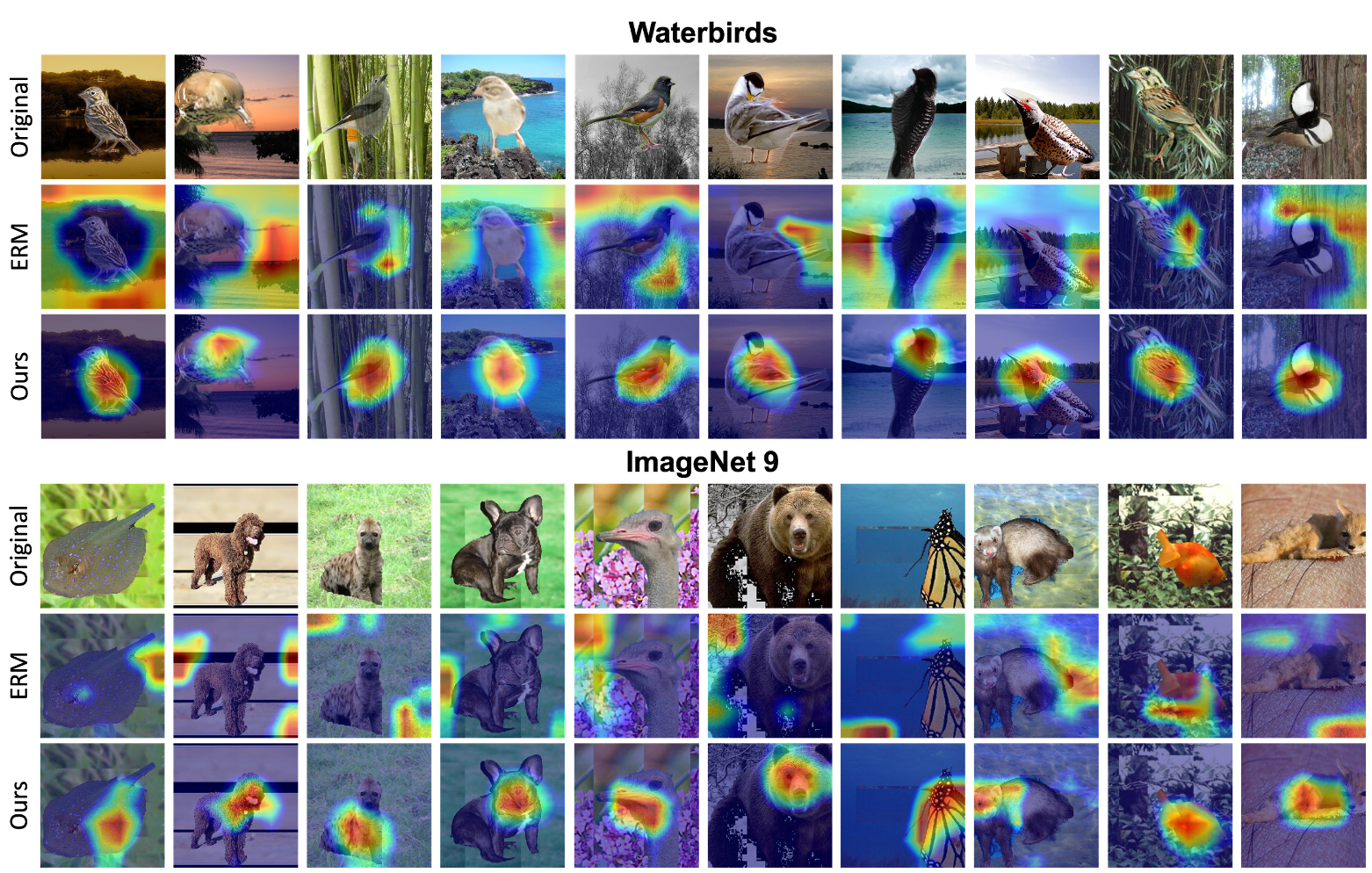}
\caption{{\small GradCAM} qualitative evaluation on Waterbirds and ImageNet-9. Dark red highlighted regions correspond to the attributions that are weighed more in the prediction. \slim~allows learning the core features instead of spuriousness.}
\label{fig:AttCompareW}
\end{figure*}

\noindent\textbf{Qualitative Evaluation of Enhanced Attention Accuracy.}
We provide more {\small GradCAM} examples showcase \slim's capability of correcting model's wrong attention in Fig.~\ref{fig:AttCompareW}.

\end{document}